\documentclass[journal]{IEEEtran}

\usepackage{hhline,bm,xcolor,amsmath,tabularx}
\usepackage{amssymb}
\usepackage{graphicx,multirow,booktabs,subfigure,comment}
\usepackage[linesnumbered,ruled]{algorithm2e}
\usepackage{caption}

\usepackage{siunitx}

\newcommand{\comp}[1]{{#1}^{\mathsf{c}}}

\def\argmin{\mathop{\rm argmin}}

\def\Var{\mathop{\rm Var}}

\usepackage{mathtools}
\DeclarePairedDelimiter\ceil{\lceil}{\rceil}
\DeclarePairedDelimiter\floor{\lfloor}{\rfloor}
\makeatletter
\newcommand*{\rom}[1]{\expandafter\@slowromancap\romannumeral #1@}
\makeatother
\newtheorem{theorem}{Theorem}
\newtheorem{lemma}[theorem]{Lemma}
\usepackage{orcidlink}
\usepackage{fixltx2e}
\usepackage[scaled=.85]{beramono}
\usepackage[T1]{fontenc}

\usepackage{lipsum}
\usepackage{fancyhdr}

\fancypagestyle{sec}{\lhead{\tmpx}}

\fancypagestyle{arXiv}
{
   \fancyhf{}
   
   \chead{\textcolor{gray}{This paper has been accepted for publication at the \\ IEEE Transactions on Neural Networks and Learning Systems,  2022\\ DOI: {10.1109/TNNLS.2022.3185742}}}
   
   \cfoot{\vspace{-6mm} \fontsize{=6pt}{10pt}\selectfont  \textcolor{gray}{© 2022 IEEE. Personal use of this material is permitted. Permission from IEEE must be obtained for all other uses, in any current or future media, including reprinting/republishing this material for advertising or promotional purposes, creating new collective works, for resale or redistribution to servers or lists, or reuse of any copyrighted component of this work in other works}\\\fontsize{=10pt}{12pt}\selectfont\thepage}
   
}

\usepackage{hyperref}

\hypersetup{
  colorlinks=false,
  linkbordercolor=white,
  urlbordercolor=white,
  pdfborder={0 0 0}
} 

\ifCLASSINFOpdf
\else
\fi
\begin{document}
%
\title{Significance Tests of Feature Relevance for \\ a Black-Box Learner \thanks{This work was supported in part by NSF DMS-1712564, DMS-1721216, and DMS-1952539, NIH grants R01GM126002, R01AG069895, R01AG065636, R01AG074858, U01AG073079, and CUHK Faulty of Science direct grant. (Corresponding author: Ben Dai.)} }
%
%
%

\author{
  Ben Dai\textsuperscript{\orcidlink{0000-0002-1620-1021}}, \IEEEmembership{Member,~IEEE,}
  Xiaotong Shen\textsuperscript{\orcidlink{0000-0003-1300-1451}},
  and Wei Pan\textsuperscript{\orcidlink{0000-0002-1159-0582}}
  \thanks{Ben Dai is with the Department of Statistics, The Chinese University of Hong Kong, Hong Kong SAR (email: bendai@cuhk.edu.hk).} 
  
  \thanks{Xiaotong Shen is with the School of Statistics, University of Minnesota, MN, 55455 USA (email: xshen@umn.edu).}

  \thanks{Wei Pan is with the Division of Biostatistics, University of Minnesota, MN, 55455 USA. (email: panxx014@umn.edu).}
}

\maketitle

\begin{abstract}
 An exciting recent development is the uptake of deep neural networks in many scientific fields, where the main objective is outcome prediction with the black-box nature. Significance testing is promising to address the black-box issue and explore novel scientific insights and interpretation of the decision-making process based on a deep learning model. However, testing for a neural network poses a challenge because of its black-box nature and unknown limiting distributions of parameter estimates while existing methods require strong assumptions or excessive computation. In this article, we derive one-split and two-split tests relaxing the assumptions and computational complexity of existing black-box tests and extending to examine the significance of a collection of features of interest in a dataset of possibly a complex type such as an image. The one-split test estimates and evaluates a black-box model based on estimation and inference subsets through sample splitting and data perturbation.  The two-split test further splits the inference subset into two but require no perturbation.
 Also, we develop their combined versions by aggregating the p-values based on repeated sample splitting.
 By deflating the \textit{bias-sd-ratio}, we establish asymptotic null distributions of the test statistics and the consistency in terms of Type \rom{2} error.
 Numerically, we demonstrate the utility of the proposed tests on seven simulated examples and six real datasets.
 Accompanying this paper is our Python library \texttt{dnn-inference} ({https://dnn-inference.readthedocs.io/en/latest/}) that implements the proposed tests.
\end{abstract}


%
\IEEEpeerreviewmaketitle

\section{Introduction}
\thispagestyle{arXiv}

%
%
%
%
\IEEEPARstart{D}{eep} neural networks \cite{schmidhuber2015deep} are a representative of black-box models, in which the learning process between features and outcomes is usually difficult to track due to the lack of knowledge about complex hidden patterns inside. The primary goal of deep learning (DL) is to fit a deep neural network for predicting outcomes with high predictive accuracy. Driven by its superior prediction performance \cite{schmidhuber2015deep}, scientists seek accountability and interpretability beyond prediction accuracy. In particular, they demand significance tests based on DL to explore novel discoveries on scientific domain knowledge, for example, if a specific lung region is significantly associated with COVID-19. 

Given a dataset, the goal of statistical significance tests is to examine if a collection of features of interest is associated with the outcome. 
A problem of this kind frequently occurs in classical parametric models or non-black-box models, for instance, in statistical genetics such as Alzheimer's disease (AD) studies, where a gene is routinely examined and tested for AD association based on a linear model. 
Yet, significance testing based on a black-box model for a more complicated dataset remains understudied. In Section \ref{sec:literature}, we discuss the existing methods and their limitations and issues. In Section \ref{sec:contribution}, we summarize our contributions to highlight the novelty of the proposed methods in addressing the existing issues.

\subsection{{Existing methods and their limitations}}
\label{sec:literature}
In the existing literature, inference methods can be categorized into two groups: non-black-box tests and black-box tests. Non-black-box tests, such as the Wald test \cite{fahrmeir2007regression} and the likelihood-ratio test \cite{king1989unifying,wasserman2020universal}, perform hypothesis testing for hypothesized features (i.e. features of interest) based on the asymptotic distribution of the estimated parameters in a parametric model such as a linear model. Black-box tests focus on a model-free hypothesis, such as Model-X knockoffs, conditional randomization tests (CRT; \cite{candes2018panning}), holdout randomization test (HRT; \cite{tansey2018holdout}), permutation test (PT; \cite{ojala2010permutation}), conditional permutation test (CPT; \cite{berrett2019conditional}), and leave-one-covariate-out test (LOCO; \cite{lei2018distribution}). Specifically, 
Model-X knockoffs conduct variable selection with FDR control based on a specified variable importance measure on each of the individual features. CRT, HRT, and CPT examine the independence between the outcome and each feature conditional on the remaining features (at least for their simulations and implementation). 
PT examines the marginal independence between the outcome and hypothesized features.
LOCO introduces the excess prediction error for each feature to measure its importance for a given dataset.

\textbf{Limitations of existing works.} {Despite the merits of the methods developed, they have their limitations. (i) First, for non-black-box tests, it is difficult to derive the asymptotic distribution of the parameter estimates from black-box models, especially for over-parametrized neural networks. Moreover, the explicit feature-parameter correspondence may be lost for a black-box model, such as a convolutional neural network (CNN; \cite{lecun1998gradient}) using shared weights for spatial pixels, and recurrent neural networks (RNN; \cite{rumelhart1986learning}) using shared weights for subsequent states. (ii) Second, most existing black-box tests focus on variable importance or inference on a {\em single feature}, yet simultaneous testing of a {\em collection} of features is more desirable in some applications. For example, in image analysis, it is more interesting to examine patterns captured by multiple pixels in a region, where the impact of every single pixel is negligible. (iii) Third, CRT, CPT, and HRT reply on a strong assumption that the conditional feature distribution is known or well-estimated, thus a test statistic can be constructed based on the generated samples from the null distribution. 
However, the complete conditionals may not be known or easy to estimate in practice, especially for complex datasets such as images or texts. 
(iv) Finally, PT, CPT, and CRT require massive computing power to refit a model many times, which is infeasible for complex deep neural networks. More detailed discussion and numerical results about the connections and differences between the existing tests and the proposed tests can be found in Sections \ref{sec:compare} and \ref{sec:num_compare}.}

\subsection{{Our contributions}}
\label{sec:contribution}

This article proposes one-split and two-split tests to address the existing issues (\textit{(i)}-\textit{(iv)}
in Section \rom{1}-A). Our main contributions are summarized as follows.

\begin{itemize}
   \item To address issues (i) and (ii), we propose a flexible \textit{risk invariance null hypothesis}
for a general loss in \eqref{eqn:loss_testing}, which measures the impact of a collection of hypothesized features
on prediction. Its relation to \textit{conditional independence} is given in Lemma \ref{lem:relation}.
   \item To address issues (iii) and (iv), the one-split and two-split tests bypass the requirement of
estimating the conditional distributions of features and testing based on the differenced empirical loss
with sample splitting subject to computational constraints.
   \item We provide theoretical guarantee of the proposed tests, c.f., Theorems \ref{thm:pOS_type1} and \ref{thm:power}, and Theorems A.1-A.4 in Appendix A. The theory is illustrated by extensive simulations.

   \item We compare the proposed tests with other existing tests and demonstrate their utility on four benchmarks on various deep neural networks in Section \ref{sec:app}.  We develop a Python library \texttt{dnn-inference} to implement the proposed tests.
\end{itemize}

Overall, the proposed tests relax the assumptions and reduce the computational cost, providing more practical,   feasible, and reliable testing for a black-box model on a complex dataset.

This article is structured as follows. Section \ref{sec:tests} introduces the one-split test as well as its combined test. Section \ref{sec:power_analysis} performs Type \rom{2} analysis and establishes the consistency of the proposed tests. Section \ref{sec:adaRatio} develops sample splitting schemes. Section \ref{sec:numerical_example} is devoted to simulation studies and applications to four real datasets. The Appendix encompasses the two-split test, additional numerical examples, and technical proofs.

\section{The proposed black-box tests}
\label{sec:tests}

In DL, a deep neural network $f(\bm{X})$ is fitted to predict an outcome $\bm Y$ based on features $\bm X \in \mathbb{R}^d$, and its prediction performance is evaluated by a loss function $l(f(\bm{X}),\bm{Y})$.
Our objective is to test the significance of a subset of features $\bm{X}_{\mathcal{S}} = \{X_j: j \in \mathcal{S} \}$ to the prediction of $\bm Y$, where $\mathcal{S}$ is an index set of hypothesized features and $\bm{X}_{\comp{\mathcal{S}}} = \{X_j: j \notin \mathcal{S}\}$ with $\comp{\mathcal{S}}$ indicating the complement set of $\mathcal{S}$. Note that $\bm{X}_{\mathcal{S}}$ can be a collection of weak features in that none of these features is jointly significant to prediction, but collectively they are. For example, in image analysis, the impact of each pixel is negligible but a pattern of a collection of pixels (e.g. in a region) may instead become significant.

To formulate the proposed hypothesis, we first generate dual data $(\bm{Z}, \bm{Y})$ by replacing $\bm{X}_{\mathcal{S}}$ by some irrelevant constants as $\bm{Z}_{\mathcal{S}}$ such as $\bm{Z}_{\mathcal{S}}=\bm 0$, that is, for $j=1, \cdots, d$,
\begin{align}
  \label{eqn:mask_feats}
     Z_j =  M, \text{ if } j \in \mathcal{S}; \quad Z_j = X_j, \textit{ otherwise},
\end{align} 
where $M$ is an arbitrary deterministic constant.

Note that the dual data $\bm{Z}$ satisfies that $\bm{Z}_\mathcal{S} \perp \bm{Y} \mid \bm{Z}_{\comp{\mathcal{S}}}$ and $\bm{Z}_{\comp{\mathcal{S}}} = \bm{X}_{\comp{\mathcal{S}}}$, thus we aim to use differences between $(\bm{X}, \bm{Y})$ and $(\bm{Z}, \bm{Y})$ to measure the impact of $\bm{X}_\mathcal{S}$ on the prediction of the outcome $\bm Y$. To proceed, we introduce the corresponding risks:
$$
R(f) = \mathbb{E}\big(l(f(\bm{X}), \bm{Y}) \big), \quad  R_\mathcal{S}(g) = \mathbb{E}\big(l(g(\bm{Z}), \bm{Y}) \big).
$$
Then, the differenced risk is defined as $R(f^*)- R_\mathcal{S}(g^*)$ to measure the significance of $\bm{X}_{\mathcal{S}}$, that is, compare the best prediction performance with/without the hypothesized features $\bm{X}_{\mathcal{S}}$ with existence of $\bm{X}_{\comp{\mathcal{S}}}$. Here $f^*=\argmin_{f} R(f)$ and $g^*=\argmin_{g} R_\mathcal{S}(g)$ are the optimal prediction functions in population.

To determine if $\bm{X}_\mathcal{S}$ is significantly relevant to the prediction of $\bm Y$, consider null $H_0$ and alternative $H_a$ hypotheses: 
\begin{equation}
\label{eqn:loss_testing}
H_0: R(f^*)- R_\mathcal{S}(g^*)=0, \quad H_a: R(f^*) - R_\mathcal{S}(g^*)<0.
\end{equation}
Rejection of $H_0$ suggests that the feature set $\bm{X}_\mathcal{S}$ is relevant to the prediction of $\bm{Y}$.
It is emphasized that, in (\ref{eqn:loss_testing}), the targets are the two true or population-level functions $f^*$ and $g^*$, instead of their estimates (based on a given sample) as implemented in some existing tests.

In the next section, we demonstrate the relation between the proposed hypothesis and the independence hypothesis. More discussion about the differences between the hypotheses in HRT and LOCO can be found in Sections \ref{sec:compare} and \ref{sec:num_compare}.

\subsection{{Connection to independence}}
This subsection illustrates the relationships among the risk invariance hypothesis in \eqref{eqn:loss_testing}, marginal independence, and conditional independence; the latter two are defined as: 
\begin{align*}
& \text{Marginal independence:} \ \bm{Y} \perp \bm{X}_\mathcal{S}, \\
& \text{conditional independence}: \ \bm{Y} \perp \bm{X}_\mathcal{S} \mid \bm{X}_{\mathcal{S}^c}.
\end{align*}
\begin{lemma}
\label{lem:relation} 
For any loss function, conditional independent implies the proposed risk invariance, that is,
$$ 
\bm{Y} \perp \bm{X}_\mathcal{S} \mid \bm{X}_{\mathcal{S}^c} \ \Longrightarrow  \ R(f^*) - R_{\mathcal{S}}(g^*) = 0.
$$
Moreover, if the negative log-likelihood or the cross-entropy $l(f(\bm{X}), Y) = - \bm{1}_{Y}^\intercal \log( f(\bm{X}) )$ is used in \eqref{eqn:loss_testing} as a loss function, then $H_0$ is equivalent to conditional independence almost surely under the marginal distribution of $\bm{X}$, that is,
\begin{align*} 
& \ R(f^*) - R_{\mathcal{S}}(g^*) = 0 \ \iff \ \text{For any } y, \\ 
& \mathbb{P} \Big( \mathbb{P} \big( Y = y \mid \bm{X}_{\mathcal{S}}, \bm{X}_{\mathcal{S}^c} \big) = \mathbb{P}\big( Y = y \mid \bm{X}_{\mathcal{S}^c} \big) \Big) = 1,
\end{align*}
\end{lemma}

As suggested by Lemma \ref{lem:relation}, conditional independence always implies risk invariance, but they can be almost surely equivalent with some particular loss functions. Hence, at any significance level, a rejection of the null hypothesis of risk invariance implies a rejection of the null hypothesis of conditional independence. Yet, such a relationship does not exist for marginal independence. Next, we present three cases with disparate loss functions to illustrate their relationships.

\noindent {\textit{Case 1.} (Constant loss): $l(f(\bm{X}), Y) = C$ for a constant $C$.}

\noindent {\textit{Case 2.} (The $L_2$-loss in regression): $l(f(\bm{X}), Y) = \mathbb{E}\big( ( Y - f(\bm{X}) )^2 \big)$ for $Y \in \mathbb{R}$.}

\noindent {\textit{Case 3.} (The cross-entropy loss in multiclass classification): $l(f(\bm{X}), Y) = -\bm{1}_{Y}^\intercal \log( f(\bm{X}))$ for $Y \in \{1, \cdots, K\}$.}

\begin{figure*}[ht]
\centering
\subfigure{\includegraphics[width=50mm]{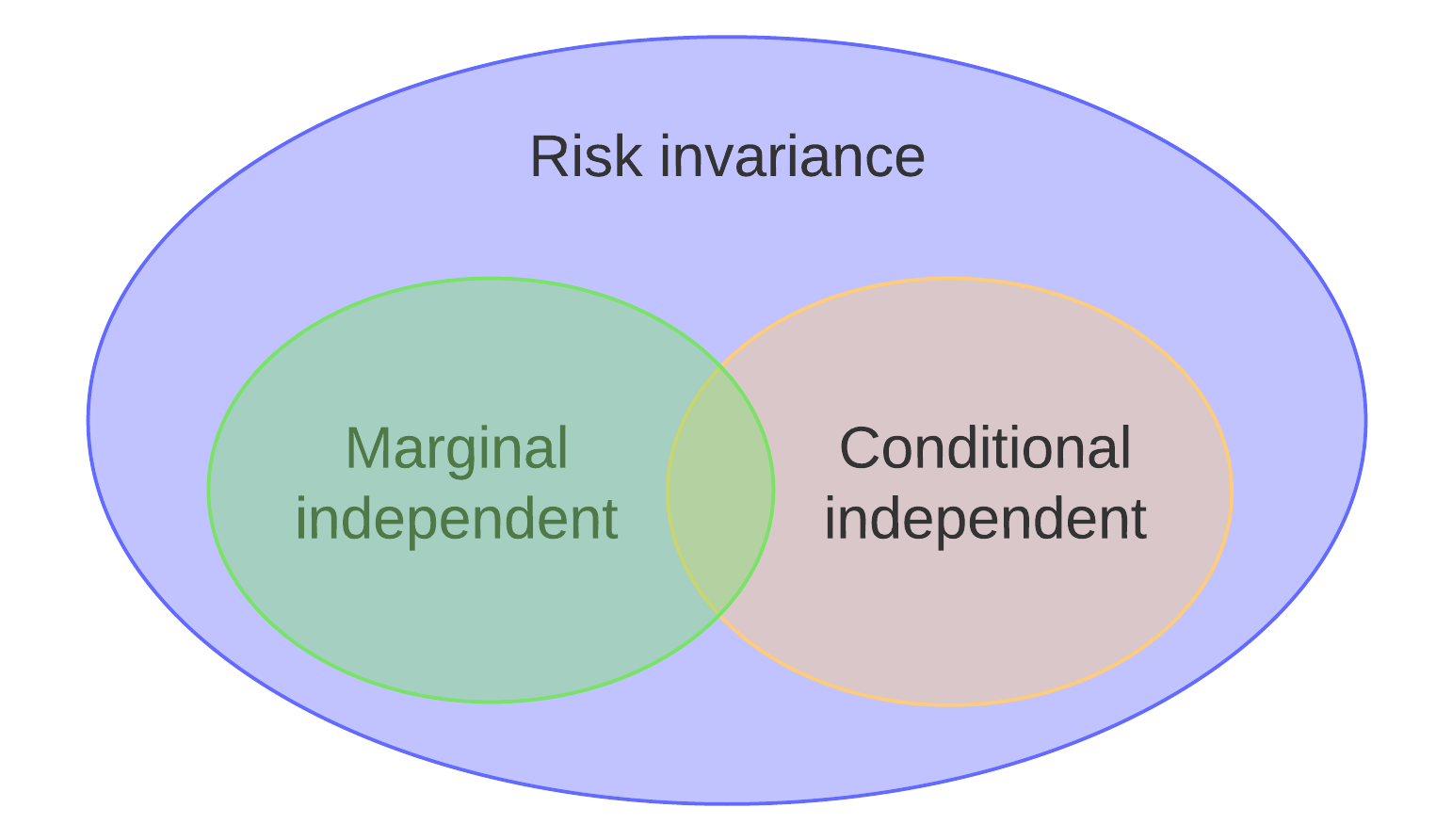}}
\subfigure{\includegraphics[width=50mm]{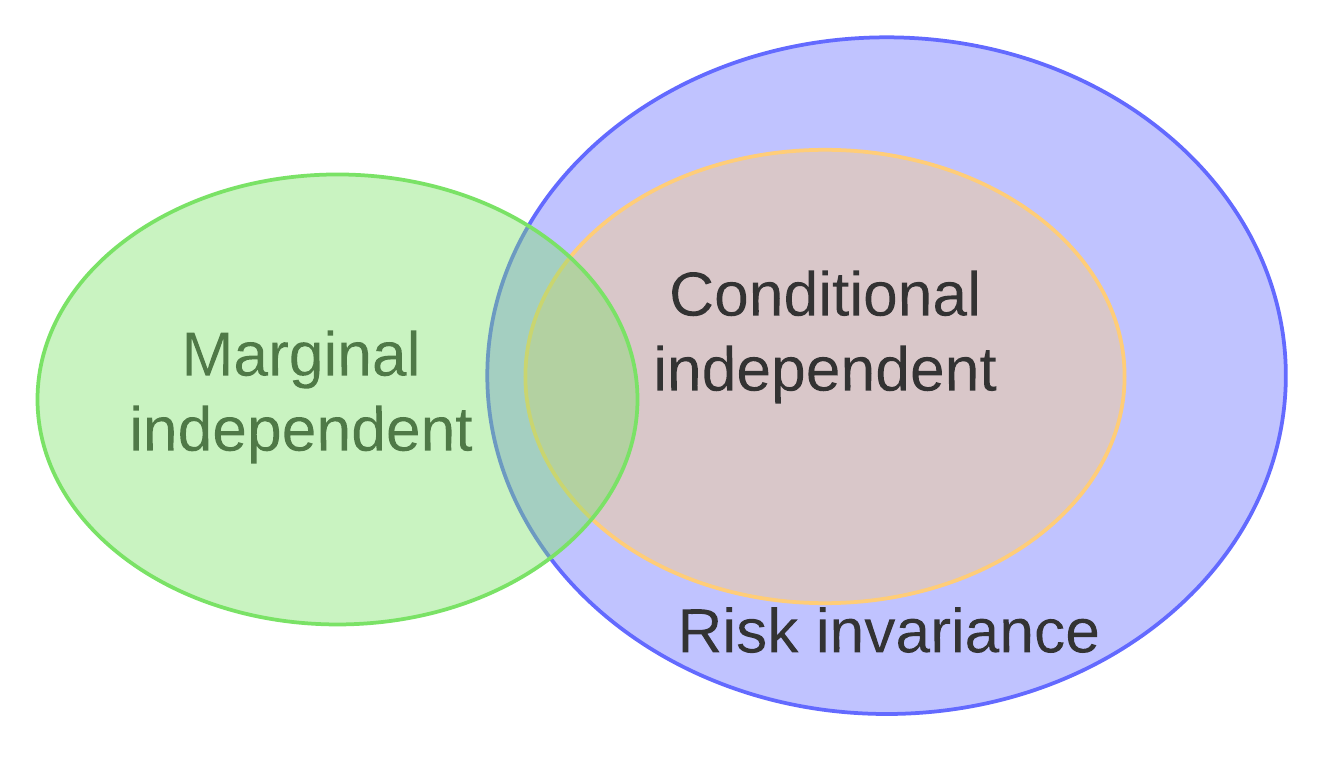}}
\subfigure{\includegraphics[width=55mm]{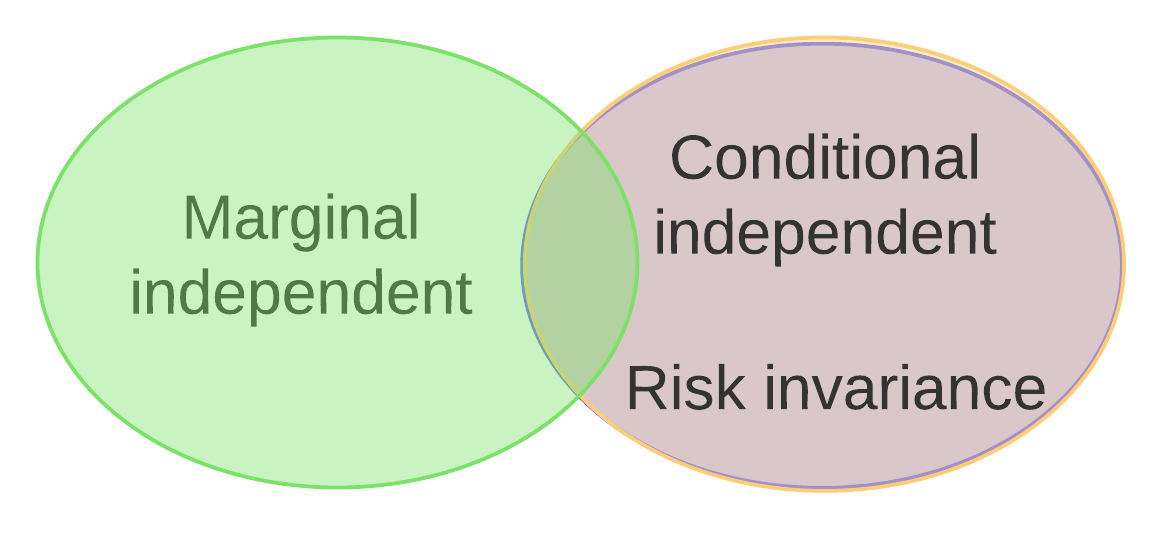}}
\caption{{Three cases illustrate different relations among marginal independence, conditional independence, and risk invariance.}}
\label{fig:venn}
\end{figure*}

As shown in Figure \ref{fig:venn},  conditional independence implies risk invariance in Cases 1 and 2 while they are equivalent in Case 3, as suggested by Lemma \ref{lem:relation}. 
In general, conditional independence or risk invariance does not yield marginal independence and vice versa.

It is worthwhile mentioning that different loss functions can lead to different conclusions. We interpret a significance test according to the loss function being used. For example, consider the misclassification error (MCE) and the cross-entropy loss for testing the relevance of $\bm{X}_\mathcal{S}$ respectively. 
With the existence of $\bm{X}_{\comp{\mathcal S}}$, the $H_0$ under the cross-entropy loss indicates that the hypothesized features are irrelevant to the conditional distribution of $\bm{Y}$ given $\bm X$, yet the $H_0$ under MCE suggests that the hypothesized features are irrelevant to classification accuracy. 

\subsection{One-split test}
\label{sec:One-split-test}
Given a dataset $\mathcal{D}_N = (\bm X_i,\bm Y_i)_{i=1, \cdots, N}$, we first split it into an estimation subset $\mathcal{E}_n = (\bm{X}_i, \bm{Y}_i)_{i=1, \cdots, n}$ and an inference subset $\mathcal{I}_m = (\bm{X}_j, \bm{Y}_j)_{j=n+1, \cdots, n+m}$, where $N$ is the number of total samples, $n = \floor{\zeta N}$ and $m = N - n$ are the sample sizes of estimation and inference subsets, and $\zeta$ is a splitting ratio. On this ground, the dual estimation subset $(\bm Z_i,\bm Y_i)_{i=1, \cdots, n}$ and the dual inference subset $(\bm{Z}_j, \bm{Y}_j)_{j=n+1, \cdots, n+m}$ can be generated based on the masking processing in \eqref{eqn:mask_feats}.
The sample splitting intends to reduce the potential bias and to prevent overfitting, especially for an over-parametrized black-box model, which  has been considered elsewhere for a different purpose in \cite{chernozhukov2018double,wasserman2009high,wasserman2020universal}. 

To access the null hypothesis in \eqref{eqn:loss_testing}, we conduct a two-level estimation to $R(f^*) - R_\mathcal{S}(g^*)$, that is, using the (dual) estimation subset to empirically estimate predictive models $(f^*, g^*)$, and the (dual) inference subset to empirically estimate two risks $R(\cdot)$ and $R_\mathcal{S}(\cdot)$. Specifically, given the (dual) estimation subset, we obtain an estimator $(\widehat{f}_n, \widehat{g}_n)$ to approximate $(f^*,g^*)$, for example, by minimizing a regularized empirical loss of a deep neural network based on the (dual) estimation subset. 
Then, the differenced empirical loss is evaluated on the (dual) inference subset based on the estimator $(\widehat{f}_n, \widehat{g}_n)$:
\begin{equation*}
  \frac{1}{m} \sum_{j=1}^m l(\widehat{f}_n(\bm{X}_{n+j}), \bm{Y}_{n+j}) -l(\widehat{g}_n(\bm{Z}_{n+j}), \bm{Y}_{n+j}).
\end{equation*}


As a remark, for flexibility, we do not specify the estimation procedure of $(\widehat{f}_n, \widehat{g}_n)$, the only explicit condition is summarized in Assumption A, which properly requires that $(\widehat{f}_n, \widehat{g}_n)$ is a consistent estimator of $(f^*, g^*)$.

One difficulty in inference is that under $H_0$ the bias of $R(\widehat{f}_n) - R_\mathcal{S}(\widehat{g}_n)$ approximating $R(f^*) - R_\mathcal{S}(g^*)$ could dominate its standard error; that is, the ratio of the bias to the standard derivation, called the \textit{bias-sd-ratio}, could be severely inflated, making the asymptotic distribution of $R(\widehat{f}_n) - R_\mathcal{S}(\widehat{g}_n)$ invalid for inference. This aspect is explained in detail in  Section \ref{sec:why_perturb}. 
To circumvent this difficulty, we present the one-split test with data perturbation to guard against the potentially inflated \textit{bias-sd-ratio} by adding an independent noise:
\begin{align}
\label{eqn:pOS_statistic}
& \Lambda^{(1)}_n = \frac{ \sum_{j=1}^{m} \Delta^{(1)}_{n,j}}{\sqrt{m}\widehat{\sigma}^{(1)}_n}, \\ 
& \Delta^{(1)}_{n,j} = l(\widehat{f}_n(\bm{X}_{n+j}), \bm{Y}_{n+j}) -l(\widehat{g}_n(\bm{Z}_{n+j}), \bm{Y}_{n+j})+\rho_n \varepsilon_j, \nonumber
\end{align}
where $\widehat{\sigma}^{(1)}_n$ is the sample standard deviation of $\{ \Delta_{n,j}^{(1)} \}_{j=1, \cdots, m}$ given $\widehat{f}_n$ and $\widehat{g}_n$:
$$
\widehat{\sigma}^{(1)}_n = \Big(\frac{1}{m-1} \sum_{j=1}^m ( \Delta_{n,j}^{(1)} - \bar{\Delta}_{n}^{(1)} )^2 \Big)^{1/2}, \quad \bar{\Delta}_{n}^{(1)} = \frac{1}{m} \Delta_{n,j}^{(1)},
$$
and $\varepsilon_j \sim N(0,1)$; $j=1, \cdots, m$ are independent noise, and $\rho_n > 0$ is the perturbation size. {Note that our proposed test is in principle similar to classical hypothesis tests using a single test statistic. For example, if we use the negative log-likelihood as the loss function, it can be regarded as an extension of the likelihood ratio test (LRT; \cite{buse1982likelihood}) to a black-box model.}
  
According to the asymptotic null distribution of $\Lambda^{(1)}_n$ in Theorem \ref{thm:pOS_type1}, we calculate the p-value $P^{(1)}=\Phi(\Lambda^{(1)}_n)$, where $\Phi(\cdot)$ is the cumulative distribution function of $N(0,1)$.

Note that $m$ is a subsequence of $n$, and $m \to \infty$ as $n \to \infty$. To derive the asymptotic null distribution of $\Lambda^{(1)}_n$, we make the following assumptions.

\noindent \textbf{Assumption A} (Estimation consistency). For some constant $\gamma > 0$, $(\widehat{f}_n, \widehat{g}_n)$ satisfies
\begin{align}
\label{assA:mean}
R(\widehat{f}_n) - R(f^*) - \big( R_\mathcal{S}(\widehat{g}_n) - R_\mathcal{S}(g^*) \big) = O_p( n^{-\gamma}), 
\end{align}
where $O_p(\cdot)$ denotes stochastic boundedness \cite{dodge2006oxford}. 

Assumption A concerns the rate of convergence in terms of the differenced regret, where $R(\widehat{f}_n) - R(f^*) \geq 0$, known as the prediction regret with respect to a loss function $l(\cdot, \cdot)$ of $\widehat{f}_n$. Note that 
\begin{align}
\label{regret}
& R(\widehat{f}_n) - R(f^*) - ( R_\mathcal{S}(\widehat{g}_n) - R_\mathcal{S}(g^*)) \nonumber \\
& \qquad \leq \max\big(R(\widehat{f}_n) - R(f^*), R_\mathcal{S}(\widehat{g}_n) -R_\mathcal{S}(g^*) \big),
\end{align}
which says that the rate $n^{-\gamma}$ is no worse than the least favorable one between the regrets of $\widehat{f}_n$ and $\widehat{g}_n$. 
In the literature, the convergence rates for the right-hand of \eqref{regret} have been extensively investigated. 
For example, the rate is $n^{-\xi /(2\xi + d)}$ for nonparametric regression \cite{wasserman2006all}, and the rate is $d n^{-2\xi /(2\xi + 1)} \log^3 n$ for a regularized ReLU neural net \cite{schmidt2020nonparametric}, where $\xi$ is the degree of smoothness of a $d$-dimensional true regression function. 
Note that an over-parametrized model may slow the convergence rate $\gamma$, yet an under-parametrized model may violate Assumption A, since the approximation error may not vanish. This fact is supported by Example 7 in Simulation (Section \ref{sec:sim}).

\noindent \textbf{Assumption B} (Lyapounov condition for $\Lambda^{(1)}_n$). Assume that 
\begin{equation*}
m^{-\mu} \mathbb{E} \big( \big|\Delta^{(1)}_{n,1} \big|^{2(1+\mu)} \big| \mathcal{E}_n \big) \stackrel{p}{\longrightarrow} 0, \quad \text{as } n \to \infty,
\end{equation*}
for some constant $\mu > 0$, where $\Delta^{(1)}_{n,1}$ is defined in \eqref{eqn:pOS_statistic}, and $\mathbb{E} \big(\cdot| \mathcal{E}_n \big)$ is the conditional expectation of inference samples given the estimation samples.

\noindent \textbf{Assumption C} (Variance condition for $\Lambda^{(1)}_n$) Assume that 
\begin{equation*}
  \Var( \Delta^{(1)}_{n,1} | \mathcal{E}_n ) \stackrel{p}{\longrightarrow} (\sigma^{(1)})^2 > 0, \quad \text{as } n \to \infty,
\end{equation*}
where $\Var( \cdot| \mathcal{E}_n )$ denotes the conditional variance of inference samples given the estimation samples.

Assumptions B and C are used in applying the central limit theorem for triangle arrays \cite{cappe2006inference}, which are verifiable under some mild conditions, c.f., Lemma C.1 in Appendix C.

The asymptotic null distribution for $\Lambda^{(1)}_n$ is indicated in Theorem \ref{thm:pOS_type1}.

\begin{theorem}[Asymptotic null distribution of $\Lambda^{(1)}_n$]
\label{thm:pOS_type1} 
In addition to Assumptions A-C, if $m = o(n^{2\gamma})$, then under $H_0$,
\begin{equation}
\label{eqn:asy_dis}
\Lambda^{(1)}_n \stackrel{d}{\longrightarrow} N(0,1), \quad \text{as} \quad n \to \infty,
\end{equation}
where $\stackrel{d}{\longrightarrow}$ denotes convergence in distribution.
\end{theorem}

Theorem \ref{thm:pOS_type1} says that the proposed test is valid under the \textit{splitting condition} of $m = o(n^{2\gamma})$. 
As a result, the estimation/inference splitting ratio needs to be suitably controlled. In Section \ref{sec:adaRatio}, we propose a ``log-ratio'' splitting scheme, in which the \textit{splitting condition} is automatically satisfied, c.f., Lemma \ref{lem:log-ratio}.

As an alternative, we present the two-split test in Appendix A to address the \textit{bias-sd-ratio} issue, where we divide inference samples further into two equal subsets for inference, in which no data perturbation is needed.

\subsection{Combining p-values over repeated random splitting}

Combining p-values via repeated random sample splitting can strengthen the one-split test \eqref{eqn:pOS_statistic}.  First, it stabilizes the testing result. Second, 
it can often empirically compensate for the power loss by combining evidence across different split samples,  as illustrated in \cite{romano2019multiple, meinshausen2009p} and our simulations in Section \ref{sec:numerical_example}. 
Subsequently, we use the order statistics of the p-values to combine the evidence from different splitting, though we could apply  other types of combining such as the corrected arithmetic and geometric means \cite{vovk2018combining,hardy1952inequalities}.


Given a splitting ratio, we repeat the random splitting scheme $U \geq 2$ times; that is, each time, we randomly split the original dataset into an estimation/inference subsets. 
In practice, $U$ cannot be large due to computational constraints and is usually 3-10 for large-scale applications. Then we compute the p-value $P^{(1)}_u$ on the $u$-th splitting; $u = 1, \cdots, U$, and combine them in two ways: the $q$-order and Hommel's weighted average of the order statistics \cite{hommel1983tests}. Specifically, 
\begin{align}
\label{eqn:cp_pvalues}
& \text{($q$-order)} \ \bar{P}^{(1)} = \min \Big(\frac{U}{q} P^{(1)}_{(q)},1\Big), \nonumber  \\
& \text{(Hommel)} \ \bar{P}^{(1)} = \min \Big(C_U \min_{1 \leq q \leq U}\frac{U}{q} 
P^{(1)}_{(q)},1\Big),
\end{align}
where $C_U=\sum_{q=1}^U \frac{1}{q}$, and $P^{(1)}_{(q)}$ is the $q$-th order statistic of $P^{(1)}_{1},\cdots, P^{(1)}_{U}$. 

The $q$-order combined test \eqref{eqn:cp_pvalues} is a generalized Bonferroni test with the Bonferroni correction factor $\frac{U}{q}$. The Hommel combined test renders robust aggregation and yields a better control of Type \rom{1} error,  where $C_U$ is a normalizing constant.


In Theorems \ref{thm:subsamples_type1} and \ref{thm:cp_power}, we further generalize the result of \cite{hommel1983tests} to control Type \rom{1} and Type \rom{2} errors of the proposed tests asymptotically. A computational scheme for the combined tests is summarized in Algorithm \ref{algo:pOS}.

\begin{theorem}[Type \rom{1} error for the combined one-split test]
\label{thm:subsamples_type1} 
Under Assumptions A-C, if $m = o(n^{2\gamma})$, then under $H_0$, for any $0<\alpha<1$ and any $U \geq 2$, the combined one-split test for \eqref{eqn:pOS_statistic} achieves
\begin{equation*}
\lim_{n \to \infty} \mathbb{P} \big(\bar{P}^{(1)} \leq \alpha | H_0 \big) \leq \alpha,
\end{equation*}
where $\bar{P}^{(1)}$ is defined in \eqref{eqn:cp_pvalues}.
\end{theorem}

\subsection{Role of data perturbation}
\label{sec:why_perturb}

This subsection discusses the role of the data perturbation for the one-split test. Now consider the one-split test without perturbation, that is, $\Lambda^{(1)}_n$ in \eqref{eqn:pOS_statistic} with $\rho_n= 0$. 
Then, we decompose $\Lambda^{(1)}_n$ into three terms:
\begin{align*}
\Lambda^{(1)}_{n} & = \frac{\sqrt{m}}{ \widehat{\sigma}^{(1)}_{n}} \Big( \frac{1}{m} \sum_{j=1}^{m} \big( \Delta^{(1)}_{n,j} - \mathbb{E}\big( \Delta^{(1)}_{n,j} | \mathcal{E}_n \big) \big) \Big) \\
& + \frac{\sqrt{m}}{\widehat{\sigma}^{(1)}_n}  \Big( R(\widehat{f}_n) - R(f^*) - \big( R_\mathcal{S}(\widehat{g}_n) - R_\mathcal{S}(g^*) \big) \Big) \nonumber \\
& \quad + \frac{\sqrt{m}}{\widehat{\sigma}^{(1)}_{n}} \big( R(f^*) - R_\mathcal{S}(g^*) \big) \equiv T_1 +T_2 + T_3.
\end{align*}
Under $H_0$, $T_3 = 0$, and $T_2$ is the \textit{bias-sd-ratio} introduced in Section \ref{sec:One-split-test}. 
Specifically, under $H_0$, as $n \rightarrow \infty$, $\widehat{\sigma}^{(1)}_{n} \stackrel{p}{\longrightarrow} 0$  as opposed to $\widehat{\sigma}^{(1)}_{n} \stackrel{p}{\longrightarrow} \sigma^{(1)} >0$ in Assumption C when $\rho_n = 0$. 
As a result, {$T_1$ may not satisfy the assumption of the central limit theorem. Furthermore, $T_2$ may not converge to zero. For example, $T_2 = O_p(m^{1/2}) \to \infty$ when $\widehat{\sigma}^{(1)}_n$ and the differenced regret $R(\widehat{f}_n) - R(f^*) - \big( R_\mathcal{S}(\widehat{g}_n) - R_\mathcal{S}(g^*) \big)$ are vanishing in the same order.} Thus, the asymptotic null distribution in \eqref{eqn:asy_dis} breaks down since $\Lambda^{(1)}_n$ is dominated by $T_2$. 

By comparison, with data perturbation, $\rho_n \rightarrow \rho > 0$, $(\sigma^{(1)})^2 = \Var \big( l(f^*(\bm{X}), \bm{Y}) - l(g^*(\bm{Z}(\bm{X})), \bm{Y}) \big)+\rho^2 > 0$. By Assumption A,
\begin{align*}
|T_2| & = \frac{\sqrt{m}}{\widehat{\sigma}^{(1)}_n} \Big| R(\widehat{f}_n) -R_\mathcal{S}(\widehat{g}_n) - \big( R(f^*) -R_\mathcal{S}(g^*) \big) \Big| \\ 
& = O_p(m^{1/2} n^{-\gamma}), \nonumber
\end{align*}
which implies that $T_2 \stackrel{p}{\longrightarrow} 0$ under the \textit{splitting condition} of  $m = o(n^{2\gamma})$. Hence, the asymptotic null distribution of $\Lambda^{(1)}_{n}$ in \eqref{eqn:asy_dis} is valid. Moreover, a ``log-ratio'' sample splitting scheme is proposed in \eqref{eqn:log-ratio}, where the \textit{splitting condition} is automatically satisfied, as indicated in Lemma \ref{lem:log-ratio}. 

In later simulations (cf. Table \ref{tab:one-vs-two}), we will show numerically that, if no data perturbation is applied in the one-split test, it leads to increasingly inflated Type I errors with larger datasets in a neural network model.

\subsection{Comparison with existing black-box tests}
\label{sec:compare}
{The one-split test in \eqref{eqn:pOS_statistic} 
has some characteristics that distinguish it from other existing black-box tests, including CRT \cite{candes2018panning}, HRT \cite{tansey2018holdout}, CPT \cite{berrett2019conditional} and LOCO tests \cite{lei2018distribution}.}


CRT, CPT, and HRT test the conditional independence of a single feature individually, and the LOCO test measures the increase in prediction error due to not using a specified feature in a given dataset.
The differences between the proposed tests and other existing tests can be summarized in three folds. 
First, for CRT, CPT, HRT, and LOCO tests, it is unclear how to test a set of multiple features $\bm{X}_\mathcal{S}$, which is the target of our tests.
Second, the significance of relevance is defined in different ways. The LOCO test conducts a significant test for the estimated model based on a given dataset with the mean absolute error, yet CRT, CPT, HRT, and the proposed tests conduct testing at the population level; that is, the former three examine conditional independence, while the last one focuses on the risk invariance as specified in \eqref{eqn:loss_testing}.
Third, CRT, CPT, and HRT require well-estimated conditional probabilities of every feature given the rest, which is often difficult in practice. Finally, the proposed tests are advantageous over CRT and CPT with reduced computational cost by avoiding a large number of model refitting.

\section{Type \rom{2} Error Analysis}
\label{sec:power_analysis}
This section performs Type \rom{2} error analysis of the one-split test  \eqref{eqn:pOS_statistic} and its combined version \eqref{eqn:cp_pvalues}. 

Consider an alternative hypothesis $H_a: R(f^*) - R_\mathcal{S}(g^*) = - m^{-1/2} \delta < 0$ for $\delta>0$. The Type \rom{2} error of the one-split test and its combined test can be written as
\begin{align*}
\beta_n(\delta) = \mathbb{P} (P^{(1)} \geq \alpha | H_a ), \quad  & \bar{\beta}_n(\delta) = \mathbb{P} ( \bar{P}^{(1)} \geq \alpha | H_a );
\end{align*}
where $\mathbb{P}(\cdot|H_a)$ denotes the probability under $H_a$, and $\alpha > 0$ is the nominal level or level of significance.

Theorems \ref{thm:power} and \ref{thm:cp_power} suggest that the one-split test and its combined test are consistent in that their asymptotic Type \rom{2} error tends to zero as $\delta \rightarrow \infty$.

\begin{theorem}[Limiting Type \rom{2} error of the one-split test]
\label{thm:power} Suppose that  the one-split test \eqref{eqn:pOS_statistic} satisfies Assumptions A-C and $m = o(n^{2\gamma})$, then
\begin{align*}
& \lim_{n \to \infty} \sup \beta_n(\delta)= \Phi\Big( z_\alpha - \frac{\delta}{\sigma^{(1)}} \Big), \ \lim_{\delta \to \infty} \lim_{n \to \infty} \sup \beta_n(\delta)= 0,
\end{align*}
where $z_\alpha = \Phi^{-1}(1-\alpha)$ is the $z$-multiplier of the standard normal distribution.
\end{theorem}

Given the results of Theorem A.3 in Appendix A, we note that the one-split test is more powerful than the two-split test in terms of the asymptotic Type \rom{2} error. 

\begin{theorem}[Limiting Type \rom{2} error of the combined tests]
\label{thm:cp_power} 
Suppose that the one-split test \eqref{eqn:pOS_statistic} satisfies Assumptions A-C and $m = o(n^{2\gamma})$,
then for $\bar{P}^{(1)}$ defined as the $q$-order combined test in \eqref{eqn:cp_pvalues}, we have
\begin{align*}
&  \lim_{n \to \infty} \sup \bar{\beta}_n(\delta) \leq \min\Big(\frac{U}{\alpha q} \Gamma,1\Big), \ \lim_{\delta \to \infty} \lim_{n \to \infty} \sup \bar{\beta}_n(\delta)= 0,  
\end{align*}
and for $\bar{P}^{(1)}$ defined as the Hommel combined test \eqref{eqn:cp_pvalues}, we have
\begin{align*}
&  \lim_{n \to \infty} \sup \bar{\beta}_n(\delta) \leq \min  \Big\{ \frac{C_U U}{\alpha q} \Gamma,1; q =1, \cdots, U\Big\}, \\ 
& \lim_{\delta \to \infty} \lim_{n \to \infty} \sup \bar{\beta}_n(\delta) = 0, 
\end{align*}
where $\Gamma = \sqrt{(\frac{q - 1}{U- q +1}) \big( \Phi ( h_0 ) - \Phi^{2}( h_0 ) - 2T(- h_0, \frac{\sqrt{3}}{3}) \big)} +  \Phi(-h_0) $, $h_0 = \frac{\delta}{ \sqrt{2} \sigma^{(1)}}$
and $T(h,a)$
is Owen's $T$ function \cite{owen1956tables}.
\end{theorem}
Note that the upper bound in Theorem \ref{thm:cp_power} can be further improved if the explicit dependency structures of the p-values from repeated sample splitting are known.

\section{Sample splitting}
\label{sec:adaRatio}

The one-split and two-split tests require the sample splitting ratio $\zeta$ to satisfy the requirement $m = o(n^{2\gamma})$ to control the Type \rom{1} error. In this section, we develop two computing schemes, namely ``log-ratio'' and ``data-adaptive'' tuning schemes, to estimate $\zeta$ in addition to the perturbation size $\rho$ for the one-split test. 

\subsection{{Log-ratio sample splitting scheme}}
This subsection proposes a log-ratio splitting scheme to ensure automatically the \textit{splitting condition} $m = o(n^{2\gamma})$. Specifically, given a sample size $N \geq N_0$, where $N_0$ is a minimal sample size required for the hypothesis testing, the estimation and inference sizes $n$ and $m$ are obtained by:
\begin{align}
\label{eqn:log-ratio}
& n = \ceil{x_0}, \quad m = N - n,  \\ 
& \text{where $x_0$ is a solution of} \quad \{x + \frac{N_0}{2 \log(N_0/2)} \log(x)  = N\}. \nonumber
\end{align}

\begin{table*}[!ht]
  \caption{Illustration of split sample sizes $(n,m)$ using the log-ratio splitting scheme \eqref{eqn:log-ratio} as the total sample size $N$ increases from $2000$ to $100000$ while $N_0=2000$ is fixed.}
  \centering
\scalebox{1.0}{
\begin{tabular}{@{}cccccccccccccccc@{}} \toprule
& ~ & \phantom{a} & & \multicolumn{9}{c}{Total sample size $(N)$} &&  \\
\cmidrule{3-15}
& ~ & \phantom{a} & 2000 & \phantom{a} & 5000 & \phantom{a} & 10000 & ~ & 20000 & ~ & 50000 & ~ & 100000\\
\midrule
& Estimation sample size $(n)$ && 1000 && 3807 && 8688 && 18578 && 48439 && 98336 \\
& Inference sample size $(m)$ && 1000 && 1193 && 1312 && 1422 && 1561 && 1664 \\
\bottomrule
\end{tabular}}
\label{tab:log-ratio}
\end{table*}

\begin{lemma}
\label{lem:log-ratio} Suppose that the estimation/inference sample sizes $(n, m)$ are determined by the log-ratio sample splitting scheme in \eqref{eqn:log-ratio}, then they satisfy the \textit{splitting condition} $m = o(n^{2\gamma})$ for any $\gamma > 0$ in Assumption A.
\end{lemma}

\subsection{Heuristic data-adaptive splitting (tuning) scheme}

The log-ratio splitting scheme in \eqref{eqn:log-ratio} is relatively conservative as the inference sample size $m$ increases in the logarithm of the estimation sample size $n$. 
To further increase a test's power, we develop a heuristic data-adaptive tuning scheme as an alternative.

The data-adaptive tuning scheme selects $(\zeta, \rho)$ by controlling the estimated Type \rom{1} error on permutation datasets. 
To proceed, we define the permutation on hypothesized features, that is, for $j = 1, \cdots, d$:
\begin{equation}
    \label{eqn:perm_feat}
     \widetilde{X}_{i,j} = X_{\pi(i),j} \text{ if } j \in \mathcal{S}, \quad \widetilde{X}_{i,j} = X_{i,j} \textit{ otherwise}, \\
\end{equation} 
where $\pi$ is a permutation mapping. Note that the hypothesized feature $\widetilde{\bm X}_{i, \mathcal{S}}$  is conditional independence to the outcome $\bm{Y}_i$ for the permutation sample. Alternatively, the null hypothesis $H_0$ is true for permutation datasets. 
On this ground, we aim to use the proportions of rejecting over $T$-times permutation as an estimate of the Type I error, and select $(\zeta, \rho)$ which is able to control the estimated Type \rom{1} error. Ideally, re-fitting and re-evaluation are required for each permutation dataset. To reduce the computational cost, we only fit $(\widehat{f}, \widehat{g})$ based on a permutation estimation subset, and estimate Type \rom{1} error by re-evaluating them at $T$-times permutation on an inference subset. The detailed procedure is summarized in following Steps 1-4.

\noindent \textit{Step 1 (Sample splitting).} Given a splitting ratio $\zeta$, split the original sample into the estimation and inference samples.

\noindent \textit{Step 2 (Permutation).} Permute hypothesized features of estimation/inference samples via \eqref{eqn:perm_feat}.

\noindent \textit{Step 3 (Fitting).} Generate the dual estimation subset via \eqref{eqn:mask_feats}, and fit $(\widehat{f}_n, \widehat{g}_n)$ based on (dual) permuted estimation subsets.

\noindent \textit{Step 4 (Estimate Type \rom{1} error).} Permute the inference subset $T$-times, and generate the corresponding dual samples via \eqref{eqn:mask_feats}. For the fixed estimators $(\widehat{f}_n, \widehat{g}_n)$, compute the (combined) p-values for each permuted (dual) inference samples under the perturbation size $\rho$, denote as $(P^{(1, t)})_{t=1, \cdots, T}$. 

Then an estimated Type \rom{1} error is computed as:
\begin{align}
\label{eqn:type1_err}
\widehat{\text{Err}^{(1)}}(\rho, \zeta) = T^{-1} \sum_{t=1}^T \mathbb{I}\big( P^{(1, t)} \leq \alpha \big). 
\end{align}

The splitting ratio $\zeta$ controls the trade-off between Type \rom{1} and Type \rom{2} errors. Specifically, a small $\zeta$ value yields biased estimators $(\widehat{f}_n,\widehat{g}_n)$, leading an inflated Type \rom{1} error, yet could reduce the Type \rom{2} error because of an enlarged inference subset.
The perturbation size $\rho$, as mentioned early, controls the \textit{bias-sd-ratio} to ensure the validity of the asymptotic null distribution. 

For the one-split test, the data-adaptive scheme estimates $(\zeta, \rho)$ as the smallest values in some candidate sets that controls the estimated Type \rom{1} error. In the process of searching candidate sets $\bm{\zeta}$ and $\bm{\rho}$, it stops once the termination criterion $\widehat{\text{Err}^{(1)}}(\rho, \zeta) \leq \alpha$ is met, which intends to reduce the computational cost. In particular, 
\begin{equation}
\label{eqn:pOS_split}
(\widehat{\rho}, \widehat{\zeta}) = \min_{\rho, \zeta} \{ \rho \in \bm{\rho}, \zeta \in \bm{\zeta} : \widehat{\text{Err}^{(1)}}(\rho, \zeta) \leq \alpha\},
\end{equation}
where $\widehat{\text{Err}^{(1)}}(\rho, \zeta)$ is the estimated Type \rom{1} error computed via \eqref{eqn:type1_err}, $\bm{\zeta}$ and $\bm{\rho}$ represent sets of candidate 
$\zeta$ and $\rho$ values.

\noindent \textbf{Overall computational cost.} {
Algorithm \ref{algo:pOS} summarizes the computational scheme of the one-split test. For the non-combining test in Algorithm \ref{algo:pOS}, the data-adaptive scheme usually requires 2-3 times of training and evaluations since the loop for the tuning of $(\zeta, \rho)$ usually terminates in one or two iterations. For the combined test, the data-adaptive scheme based on 5 random splits usually requires 10 times of training and evaluations. 
The running time for the proposed test is indicated in Tables \ref{tab:sim_sample_size} and B.1 in Appendix B.}

\begin{algorithm}
    \SetKwInOut{Input}{Input}
    \SetKwInOut{Output}{Output}
    \Input{{Data: $(\bm{x}_i, \bm{y}_i)_{i=1}^N$; Set of hypothesized features: $\mathcal{S}$; Number of splitting: $U$}}
    \Output{{p-value for testing \eqref{eqn:loss_testing}}}
    Estimate $(\widehat{\rho}, \widehat{\zeta})$ from \eqref{eqn:pOS_split} \;
    \For{$u = 1, \cdots, U$}
      {
      Shuffle the data\;
      Split the data into an estimation subset and an inference subset, where $m = \hat{\zeta}N$ and $n = N -m$\;
      Generate dual samples via \eqref{eqn:mask_feats} for estimation/inference subsets\;
      Compute $\Lambda^{(1)}_u$ from \eqref{eqn:pOS_statistic} with $\widehat{\rho}$\;
      Compute p-value $P^{(1)}_{u} = \Phi(\Lambda^{(1)}_u)$
      } 
  \uIf(\tcp*[f]{combined one-split test}){$U > 1$}
   { Compute the combined p-value $\bar{P}^{(1)}$ via \eqref{eqn:cp_pvalues} \;
   \Return{p-value $\bar{P}^{(1)}$}
   }
  \Else(\tcp*[f]{non-combined one-split test}){
  \Return{p-value $P^{(1)}_1$}}
    \caption{One-split test for region significance}
    \label{algo:pOS}
\end{algorithm}

\section{Numerical examples}
\label{sec:numerical_example}

This section examines the proposed tests for their capability of controlling Type \rom{1} and Type \rom{2} errors in both simulated and real examples. All tests are implemented in our Python library \texttt{dnn-inference} (\url{https://github.com/statmlben/dnn-inference}).

\subsection{{Numerical comparison with existing black-box tests}}
\label{sec:num_compare}
{
This subsection presents a simple example to illustrate the differences between the proposed tests and other existing black-box tests, including the holdout randomization test (HRT; \cite{tansey2018holdout}), the leave-one-covariate-out test (LOCO; \cite{lei2018distribution}), the permutation test (PT; \cite{breiman2001random,ojala2010permutation}), and the holdout permutation test (HPT; \cite{tansey2018holdout}).
For PT, we use the scheme of \cite{ojala2010permutation} to permute multiple hypothesized features $\bm{X}_{\mathcal{S}}$, on which we refit the model, and the permutation size is 100. Algorithm 2 in Appendix B summarizes the procedure for the permutation test.
Note that we exclude CRT here due to its enormously expensive computing in refitting a model many times.}

{
To alleviate the high computational cost of refitting, HPT uses data-splitting into a training sample and a test sample. Then it fits only one time on training data and performs the permutation test over the test sample with the trained model. 
In our context, we extend HPT in \cite{tansey2018holdout} by simultaneously permuting multiple hypothesized features $\bm{X}_{\mathcal{S}}$.}

{
One issue with the PT and HPT is that permutations of hypothesized features usually alter the dependence structure between $\bm{X}_{\mathcal{S}}$ and $\bm{X}_{\comp{\mathcal{S}}}$.  As a result,   the sampling distribution based on permuted samples may differ from the null distribution. For example, the simulated example in Appendix B.2 indicates that both HPT and PT lead to dramatically inflated Type \rom{1} errors.  }





\begin{table*}[!ht]
  \caption{{Returning values of the one-split/two-split tests and other existing black-box tests. Here one-split, two-split, HRT, LOCO, PT, and HPT denote the proposed tests in Algorithms \ref{algo:pOS} and Algorithm 1 in Appendix A, holdout randomization test (HRT; \cite{tansey2018holdout}), the leave-one-covariate-out test (LOCO; \cite{lei2018distribution}), permutation test (PT), and holdout permutation test (HPT; \cite{tansey2018holdout}).}}
  \centering
\scalebox{1.}{
\begin{tabular}{@{}cccccp{5.5cm}cccccccccc@{}} \toprule
& Test & \phantom{a} & Return & \phantom{a} & $H_0$ & \phantom{a} & ~  \\
\midrule
& One-split && p-value && risk-invariance $R(f^*) = R_\mathcal{S}(g^*)$, && 0.003 \\
& Two-split && p-value && risk-invariance $R(f^*) = R_\mathcal{S}(g^*)$ && 0.018 \\
& HRT && p-values for all feats && conditional indep $\bm{X}_{j} \perp \bm{Y} | \bm{X}_{-j}$ && (0.840, 0.045, 0.064, 0.900, 0.158)  \\
& LOCO && p-values for all feats && equal errors with/without feat $j$ for a given dataset && (0.132, 0.791, 0.180, 0.435, 0.342) \\
& PT && p-value && marginal indep $\bm{X}_{\mathcal{S}} \perp \bm{Y}$ && 0.010 \\
& HPT && p-value && marginal indep $\bm{X}_{\mathcal{S}} \perp \bm{Y}$ && 0.001 \\
\bottomrule
\end{tabular}}
\label{tab:diff}
\end{table*}

In this section, we generate a random sample of size $N=1000$. First, $\bm{X}=(X_1, \cdots, X_5)^\intercal$ follows a uniform 
distribution on $[-1,1]$ with a pairwise correlation $\rho_{ij} = 0.5^{|i-j|}$; $i,j = 1,\cdots, 5$. Second,  the outcome $Y$ is generated 
as $Y = 0.02(X_1 + X_2 + X_3) + 0.05 \epsilon$, where $\epsilon \sim N(0,1)$.

{A simulation study is performed for the one-split and two-split tests, HRT, LOCO, and HPT. For HRT, we use
the code of \cite{tansey2018holdout} available at GitHub with a default mixture density network with 2 components. For other methods, we fit a linear function based on stochastic gradient descent (SGD) with the same fitting parameters, that is, \texttt{epochs} is 100, \texttt{batch\_size} is 32, and early stopping with \texttt{validation\_split} being 0.2 and \texttt{patience} being 10, where \texttt{patience} is the number of epochs until termination if no progress is made on the validation set. For HRT, LOCO, HPT, and PT, the sample splitting ratio is fixed as 0.8, and the data-adaptive scheme is used for the proposed tests.}

{The returning values are summarized in Table \ref{tab:diff}: the one-split and two-split tests return valid p-values for the hypothesis in \eqref{eqn:loss_testing} with $\mathcal{S} = \{1,2,3\}$, HRT and LOCO return p-values for individual features of conditional independence and error-invariance for a given dataset, respectively. PT and HPT provide p-values for marginal independence. Therefore, the proposed tests are the only ones targeting the specified null hypothesis in \eqref{eqn:loss_testing}.}

\subsection{Simulations}
\label{sec:sim}
Consider a nonparametric regression model,
\begin{equation}
\label{eqn:network_example}
Y = f^*(\bm{X})+ \epsilon, \quad \epsilon \sim N(0, 1),
\end{equation}
where $f^*(\bm x)$ is an unknown function on $\bm{x} \in [-1,1]^d$. It is known that $f^*(\bm{x})=g^*(\bm z)$ only depends on a subset of features of $\bm x$, in which $\bm{z}_{\mathcal{S}_0}=\bm{0}$ and $\bm{z}_{\mathcal{S}^c_0} = \bm{x}_{\mathcal{S}^c_0}$ with $\mathcal{S}_0 = \{ 1, \cdots, |\mathcal{S}_0| \}$. Given a hypothesized index set $\mathcal{S}$, our goal is to test if $\bm{X}_{\mathcal{S}}$ is relevant to predicting the outcome $Y$, as specified in \eqref{eqn:loss_testing}.

For illustration, we set the regression function as a neural network $f^*(\bm{x}) = A 
\big( \bm{W}_{L}^* A \big( \bm{W}_{L-1}^* \cdots A(\bm{W}_{1}^* \bm{x}) \big) \big)$, where $A(\cdot)$ is the ReLU activation function, $\bm{W}_{l}^* = (w_{l,ij}^*) \in \mathbb{R}^{d_{l} \times d_{l-1}}$ is a weight matrix, $\| \bm{w}_{l,j}^* \|_2 = \tau/d_{l-1}^{1/2}$, $\bm{w}_{l,j}^*$ is the $j$-th column of the matrix $\bm{W}_l^*$, $\tau > 0$ is a constant, $d_l$ is the width for the $l$-th layer, and $d_0 = d, \ d_L = 1, \ d_1 = \cdots = d_{L-1} = \varpi$ and $L$ is the depth of the network.
Clearly, $f^* \in \mathcal H$, where $\mathcal{H}$ is a candidate class defined as:
\begin{align*}
  \mathcal{H} & = \{ f(\bm x)=A \big( \bm{W}_{L} A \big(\bm{W}_{L-1} \cdots A(\bm{W}_{1} \bm{x}) \big) \big): \\
  & \hspace{3.5cm} \|\bm{W}_{l}\|_2 \leq \tau, \| \bm{W}_{l} \|_{2,1} \leq \tau \}.
\end{align*}
We perform simulations under the model in \eqref{eqn:network_example}, where $\bm{X} \sim N(\bm{0}, B \bm{\Sigma})$, $\bm{\Sigma}_{ij} = r^{|i-j|}$, $r \in [0,1)$, $r$ represents the correlation coefficient of features, $B$ controls the magnitude of the features, $(L, \varpi, \tau)$ denotes the depth, width, and the $L_2$-norm of the neural network, {$\mathcal{S}_0 = \{1, \cdots, |\mathcal{S}_0|\}$ is an index set of the true non-discriminative features, and $\mathcal{S}$ is an index set of hypothesized features.}

For hypotheses in \eqref{eqn:loss_testing}, we examine four index sets of hypothesized features $\mathcal{S}$:

(i) $\mathcal{S}=\{1, \cdots, |\mathcal{S}_0| \}$, 

(ii) $\mathcal{S} = \{ \floor{|\mathcal{S}_0|/2}, \cdots, \floor{|\mathcal{S}_0|/2} + |\mathcal{S}_0| \}$, 

(iii) $\mathcal{S} = \{ \floor{p/2}, \cdots, \floor{p/2} + |\mathcal{S}_0| \}$, 

(iv) $\mathcal{S} = \{ p - |\mathcal{S}_0|, \cdots, p \}$. 

\begin{figure*}[ht]
\centering
\includegraphics[scale=.15]{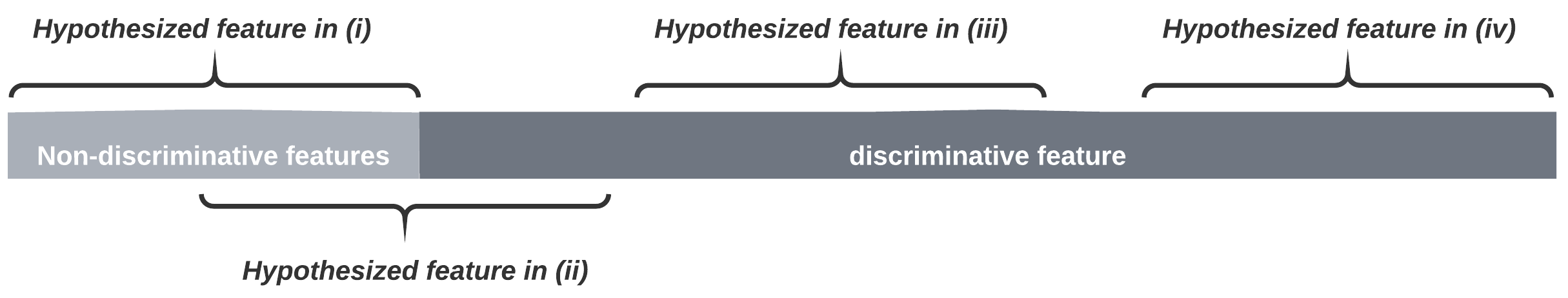}
\caption{{Illustration of four index sets of hypothesized features in simulations: (i) Type I error analysis, (ii)-(iv): Type \rom{2} error analysis. Note that the impact of the hypothesized features $\mathcal{S}$ on $\mathcal{S}_0$ decreases while the Type \rom{2} error is expected to decrease from (ii) to (iv).}}
\label{fig:sim_sets}
\end{figure*}

These four sets are illustrated in Figure \ref{fig:sim_sets}.
Note that $\mathcal{S} \bigcup \mathcal{S}_0 = \mathcal{S}_0$ in (i), implying that it is for Type \rom{1} error analysis, while $\mathcal{S} \bigcup \mathcal{S}_0 \neq \mathcal{S}_0$ in (ii)-(iv), suggesting Type \rom{2} error analysis. 
From (ii) to (iv), the distance (or correlation) between the hypothesized features $\mathcal{S}$ and those non-discriminative features in $\mathcal{S}_0$ is increasing (or decreasing), thus the Type \rom{2} error is expected to go down. Seven examples are considered for (i)-(iv) hypothesized feature sets.

\textbf{Example 1.} (\textit{Impact of the sample size and splitting method})
This example (Table \ref{tab:sim_sample_size}) concerns the performance of the proposed tests in relation to the sample size $N$ based on \emph{log-ratio} and \emph{data-adaptive} splitting methods, where $N$ ranges from 2000 to 10000, $B=0.4$, $r = 0.25$, $p=100$, $\varpi = 128$, $L=3$, $|\mathcal{S}_0|=5$.

\textbf{Example 2.} (\textit{Impact of the strength of features of interest}) This example (Table \ref{tab:sim_B}) concerns the performance of the proposed tests with respect to the magnitude of features $B$, where $B = 0.2, 0.4, 0.6$, $N=6000$, $p=100$, $r=0.25$, $\varpi = 128$, $L=3$, and $|\mathcal{S}_0|=5$. The data-adaptive tuning scheme is applied for this example.

\textbf{Example 3.} (\textit{Impact of the depth and width of a neural network}) This example (Table B.1 in Appendix B) concerns the performance of the proposed tests in terms of the width $\varpi$ and depth $L$ of a neural network, where $N=6000$, $L = 2, 3, 4$, $\varpi = 32, 64, 128$, $B=0.4$, $r=0.25$, $p=100$ and $|\mathcal{S}_0|=5$.

\textbf{Example 4.} (\textit{Impact of the number of hypothesized features}) This example (Table B.2 in Appendix B) concerns the proposed tests with respect to the number of hypothesized features $|\mathcal{S}_0|$, where $|\mathcal{S}_0| = 5, 10, 15$, $N=6000$, $B=0.4$, $p=100$, $\varpi = 128$, $r = 0.25$, and $L=3$.

\textbf{Example 5.} (\textit{Impact of feature correlations}) This example (Table B.3 in Appendix B) concerns the proposed tests in terms of the feature correlation $r$, where $r = 0.00, 0.25, 0.50$, $N=6000$, $B=0.4$, $p=100$, $\varpi = 128$, $L=3$.

\textbf{Example 6.} (\textit{Impact of different modes of combining p-values}) This example (Table B.4 in Appendix B) concerns the combined tests with different ways of combining p-values. Type \rom{1}/\rom{2} errors are examined in two simulated examples: (1) $N=6000, \rho=0.25, B=0.2, L=3, \varpi=128$; (2) $N=6000, \rho=0.25, B=0.4, L=4, \varpi=32$.

\textbf{Example 7.} (\textit{Impact of over/under-parameterized models}) This example (Table \ref{tab:sim_diff}) concerns the impact of the proposed tests based on different underlying black-box models. Specifically, we set the ground truth function $f^*$ as a neural network with $\varpi = 64$ and $L=3$, and consider both the under-parameterized and over-parameterized models with $\varpi = 32, 64, 128$.

\begin{table*}[!ht]
  \caption{Empirical Type \rom{1}/\rom{2} errors of the (combined) one-/two-split tests, and their combined tests in Example 1 at $\alpha = 0.05$. }
  \centering
\scalebox{0.9}{
\begin{tabular}{@{}cccccccccccccccc@{}}
\toprule
& Splitting method & \phantom{a} & Test & \phantom{a} & Sample size & \phantom{a} & Type \rom{1} error & \phantom{a} & Type \rom{2} error & \phantom{a} & Time (Second) & \\
\midrule
& \emph{Log-ratio} && One-split && 2000 && 0.004 && (0.78, 0.12, 0.08) && 8.1(0.4) \\
& && &&  6000 && 0.004 && (0.58, 0.00, 0.00) && 9.6(0.6) \\
& && &&  10000 && 0.010 && (0.55, 0.00, 0.00) && 11.7(0.4) \\
\cmidrule{4-12}
& && Two-split && 2000 && 0.026 && (0.89, 0.66, 0.65) && 8.4(0.4) \\
& && & & 6000 && 0.036 && (0.91, 0.55, 0.58) && 9.7(0.5) \\
& && & & 10000 && 0.034 && (0.84, 0.54, 0.57) && 11.4(0.2) \\
\cmidrule{4-12}
&& & Comb. one-split && 2000 && 0.016 && (0.76, 0.05, 0.05) && 42.1(1.6) \\
& && && 6000 && 0.012 && (0.49, 0.00, 0.00) && 45.6(1.3) \\
& && && 10000 && 0.010 && (0.33, 0.00, 0.00) && 56.3(0.8) \\
\cmidrule{4-12}
& && Comb. two-split && 2000 && 0.018 && (0.90, 0.70, 0.68) && 40.8(1.6) \\
& && && 6000  && 0.024 && (0.91, 0.51, 0.53) && 45.5(1.2) \\
& && && 10000 && 0.018 && (0.92, 0.59, 0.58) && 56.5(1.1) \\
\toprule
& \emph{Data-adaptive} && One-split && 2000 && 0.043 && (0.75, 0.21, 0.15) && 15.2(0.1) \\
& &&  &&  6000 && 0.050 && (0.39, 0.01, 0.00) && 41.2(0.3) \\
& &&  &&  10000 && 0.049 && (0.11, 0.00, 0.00) && 66.0(0.4) \\
\cmidrule{4-12}
& && Two-split && 2000 && 0.050 && (0.89, 0.74, 0.69) && 14.0(0.1) \\
& & && & 6000 && 0.035 && (0.82, 0.49, 0.42) && 37.0(0.2) \\
& & && & 10000 && 0.040 && (0.81, 0.23, 0.25) && 61.6(0.4) \\
\cmidrule{4-12}
& && Comb. one-split && 2000 && 0.034 && (0.74, 0.00, 0.05) && 37.9(0.1) \\
& && && 6000 && 0.046 && (0.14, 0.00, 0.00) && 68.3(0.3) \\
& && && 10000 && 0.045 && (0.00, 0.00, 0.00) && 107.2(0.7) \\
\cmidrule{4-12}
& && Comb. two-split && 2000 && 0.015 && (0.91, 0.74, 0.71) && 38.0(0.1) \\
& && && 6000  && 0.030 && (0.90, 0.30, 0.35) && 76.3(0.5) \\
& && && 10000 && 0.014 && (0.87, 0.07, 0.08) && 110.3(0.5)  \\
\bottomrule
\end{tabular}}
\label{tab:sim_sample_size}
\end{table*}

\begin{table}[!ht]\centering
  \caption{Empirical Type \rom{1}/\rom{2} errors of the (combined) one-split and two-split tests in Example 2.}
\scalebox{1.0}{
\begin{tabular}{@{}cccccccccccccccc@{}} \toprule
& Test & \phantom{a} & $B$ & \phantom{a} & Type \rom{1} error & \phantom{a} & \multicolumn{1}{c}{Type \rom{2} error} \\
\midrule
& One-split && 0.2 && 0.057 && (0.76, 0.32, 0.12) \\
&& & 0.4 && 0.050 && (0.29, 0.01, 0.00) \\
&& & 0.6 && 0.057 && (0.03, 0.00, 0.00) \\
\midrule
& Two-split && 0.2 && 0.049 && (0.94, 0.88, 0.86)  \\
&& &  0.4 && 0.035 && (0.82, 0.49, 0.42) \\
&& &  0.6 && 0.041 && (0.63, 0.03, 0.02) \\
\midrule
& Comb. one-split && 0.2 && 0.027 && (0.73, 0.07, 0.07) \\
&& &  0.4 && 0.046 && (0.14, 0.00, 0.00)  \\
&& &  0.6 && 0.033 && (0.00, 0.00, 0.00) \\
\midrule
& Comb. two-split && 0.2 && 0.019 && (1.00, 1.00, 0.97) \\
&& &  0.4 && 0.030 && (0.90, 0.30, 0.35)  \\
&& &  0.6 && 0.012 && (0.55, 0.00, 0.00) \\
\bottomrule
\end{tabular}}
\label{tab:sim_B}
\end{table}

  \begin{table}[!ht]\centering
    \caption{Empirical Types \rom{1}/\rom{2} errors of the (combined) one-/two-split tests in Example 7 based on $\varpi_0=64$ (width for the truth model) and different $\varpi$s (width for a learning model).}
  \scalebox{1.0}{
  \begin{tabular}{@{}cccccccccccccccc@{}} \toprule
  & Test & \phantom{a} & $\varpi$ & \phantom{a} & Type \rom{1} error & \phantom{a} & \multicolumn{1}{c}{Type \rom{2} error} \\
  \midrule
  & One-split && 32 && 0.067 && (0.80, 0.36, 0.37) \\
  && & 64 && 0.025 && (0.81, 0.32, 0.29) \\
  && & 128 && 0.017 && (0.80, 0.28, 0.26) \\
  \midrule
  & Two-split && 32 && 0.017 && (0.97, 0.94, 0.93) \\
  && &  64 && 0.020 && (0.97, 0.94, 0.93) \\
  && &  128 && 0.033 && (0.96, 0.93, 0.93) \\
  \midrule
  & Comb. one-split && 32 && 0.140 && (0.58, 0.16, 0.11) \\
  && &  64 &&  0.030 && (0.81, 0.17, 0.14) \\
  && &  128 && 0.013 && (0.85, 0.18, 0.16) \\
  \midrule
  & Comb. two-split && 32 && 0.013 && (0.96, 0.93, 0.91) \\
  && &  64 &&  0.027 && (0.96, 0.93, 0.94) \\
  && &  128 && 0.007 && (0.97, 0.95, 0.97) \\
  \bottomrule
  \end{tabular}}
  \label{tab:sim_diff}
  \end{table}


For a test's Type \rom{1} and \rom{2} errors, we compute the proportions of its rejecting $H_0$ out of 1000 simulations under $H_0$ and out of 100 simulations under $H_a$, respectively.

When implementing the log-ratio splitting scheme, $(n,m)$ is determined by \eqref{eqn:log-ratio} with $N_0 = 1000$, and $\rho = 0.01$; for the data-adaptive scheme, the grids of $\zeta$ are set as $\{0.2, 0.4, 0.6, 0.8\}$. Moreover, the grids for searching the optimal perturbation size are $\{0.01, 0.05, 0.1, 0.5, 1.0\}$. For combined tests, the number of repeated random splitting is set as 5. The hyperparameters of fitting a neural network are the same as in Section \ref{sec:num_compare}.


\textbf{Type \rom{1}/\rom{2} errors of the (combined) one-split/two-split tests.} As indicated in Tables \ref{tab:sim_sample_size}-\ref{tab:sim_B} and Tables B.1 - B.5, the one-split/two-split tests perform well in all examples with respect to controlling Type \rom{1}/\rom{2} errors. 
In particular, Type \rom{1} errors are close to the nominal level $\alpha = 0.05$, whereas Type \rom{2} errors decrease to 0 as the sample size $N$ increases. As expected, the one-split test outperforms the two-split test in terms of Type \rom{2} error, which agrees with Theorems \ref{thm:power} and Theorem A.3 in Appendix A. 
The combined tests consistently improve the performance in terms of both Type \rom{1}/\rom{2} errors.

\textbf{Runtime.}  The combined tests may double the runtime of their non-combined counterparts based on the data-adaptive tuning scheme. 
This result suggests that the one-split/two-split, and their combined tests are practically feasible for black-box testing subject to computational constraints as in the case of applying deep neural networks to large data.

\textbf{Combining p-values.} As suggested by Table B.4 in Appendix, the Hommel combining method controls the Type \rom{1} error while having reasonably good power in reducing the Type \rom{2} error. The Bonferroni and Cauchy methods have an issue of failing to control Type \rom{1} error, whereas other combining methods are conservative in the first case of Example 6.

\textbf{Over/under-parameterized models.} As suggested in Table \ref{tab:sim_diff}, an under-parameterized model $(\varpi=32)$ has inflated Type \rom{1} errors, which agrees with the theoretical analysis in Section \ref{sec:One-split-test}; an over-parameterized model $(\varpi = 128)$ is able to control the Type \rom{1} error, and provides similar performance in the power or Type \rom{2} error to that of  the perfectly specified model (with exactly the same network structure of the ground truth model), it is partially because the early-stopping is conducted as a regularization for over-parameterized models. For the (combined) two-split test, both the over- and under-parameterized models perform similarly to the perfectly specified model. One plausible explanation (for its no inflation of Type \rom{1} errors) is that the two-split test is conservative in the finite-sample setting.

We summarize the advantages of the different tests and the combining/tuning methods in Table \ref{tab:adv}.

\begin{table*}[ht]
  \caption{{Advantage for different tests, combining, and tuning methods.}}
  \centering
\begin{tabular}{llllp{6cm}lp{5cm}}
\toprule
&& && \textit{Advantage} && Evidence  \\
\midrule
\multirow{2}{*}{Test} && One-split && \textit{More powerful} && Tables \ref{tab:sim_sample_size}, \ref{tab:sim_B}, Tables B.1-B.4 in Appendix B   \\
&& Two-split && \textit{No need to perturb data} && Appendix A \\
\midrule
\multirow{2}{*}{Combine} && Comb. && \textit{More powerful} && Tables \ref{tab:sim_sample_size}, \ref{tab:sim_B}, Tables B.1-B.4 in Appendix B \\
&& Non-comb. && \textit{Less computation time} && Table \ref{tab:sim_sample_size} \\
\midrule
\multirow{2}{*}{Ratio} && Data-adaptive && \textit{More powerful} && Tables \ref{tab:sim_sample_size}, \ref{tab:sim_B}, Tables B.1-B.4 in Appendix B \\
&& Log-ratio && \textit{No need to tune the ratio, and less computation time} && Lemma \ref{lem:log-ratio}, Table \ref{tab:sim_sample_size}\\
\bottomrule
\end{tabular}
\label{tab:adv}
\end{table*}

\subsection{One-split test and perturbation}
Consider a regression model in \eqref{eqn:network_example}, where $\mathcal{S}_0 = \{1, 2, 3\}$, $\bm{X} \sim N(\bm{0}, B \bm{\Sigma})$, where $\Sigma_{1j} = \Sigma_{j1} = 0.1$; $j = 1, \cdots, p$, and $\Sigma_{ij} = 0$, if $i,j \neq 1$ and $i \neq j$. In this case, let $\mathcal{S} = \mathcal{S}_0$, then $H_0$ is true in the population level. 
Furthermore, only partial features are observed in a dataset $(\bm{x}^{(N)}_i, y^{(N)}_i )_{i=1}^N$, where $\bm{x}^{(N)}_i = (\bm{x}_{i1}, \cdots, \bm{x}_{id_N})^\intercal$ and $y^{(N)}_i$ is generated as $y^{(N)}_i = f^*(\tilde{\bm{x}}^{(N)}_i) + \epsilon_i$, $d_N \leq d$ is the number of observed features and $d_N \to d$ as $N \rightarrow \infty$, and $\tilde{\bm{x}}_i^{(N)} = (\bm{x}_{i1}, \cdots, \bm{x}_{id_N}, 0, \cdots, 0)^\intercal$ is a $d$-dimensional dummy variable.

Then, we simulate a dataset $(\bm{x}^{(N)}_i, y^{(N)}_i)_{i=1}^N$, with $d=100$, $d_N = \floor{d(1 - 1/ \log(N) )}$, and $N = 2000, 6000, 10000$. For implementation, we set $\zeta=0.2$ for the one-split and two-split tests and  $\rho=1.0$ for the one-split test. {The fitting parameters of a neural network remain the same as in Section \ref{sec:sim}.
Then, the Type \rom{1} errors based on the two-split test, and the one-split tests with/without perturbation are reported in Table \ref{tab:one-vs-two}.}
\begin{table}[!ht]
  \caption{Type \rom{1} errors of the one-split tests with/
  without perturbation (PTB) and the two-split test in Section 6.4.}
  \centering
\scalebox{1.0}{
\begin{tabular}{@{}cp{1.6cm}ccccccccccccccccc@{}} \toprule
\phantom{a} & Test & \phantom{a} & $N=2000$ & \phantom{a} & $N=6000$ & \phantom{a}& $N=10000$  \\
\midrule
& One-split \text{without PTB} && 0.083 && 0.109 && 0.193 \\
\cmidrule{1-8}
& One-split \text{with PTB} && 0.057 && 0.053 && 0.061 \\
\cmidrule{1-8}
& Two-split && 0.048 && 0.051 && 0.047 \\
\bottomrule
\end{tabular}}
\label{tab:one-vs-two}
\end{table}


As indicated in Table \ref{tab:one-vs-two}, the two-split test and the one-split with perturbation approximately control Type \rom{1} errors across all situations, whereas the one-split test without perturbation has inflated Type \rom{1} errors significantly exceeding the nominal level $\alpha=0.05$.

\section{Real Application}
\label{sec:app}
\subsection{MNIST handwritten digits}
\label{sec:app_mnist}

This subsection applies the proposed test to the MNIST handwritten digits dataset \cite{lecun1998gradient}. The MNIST dataset is a standard benchmark for XAI methods \cite{ribeiro2016should}, in part because the results of detection could be easily evaluated by human visual intuition.  In particular, we extract $14,251$ images from the dataset with labels `7' and `9' to discriminate between these two digits. Our primary goal is to test certain image features differentiating digit '7' from digit '9', where a marked region of an image specifies hypothesized features.

\begin{figure}[ht]
\centering
\includegraphics[scale=.26]{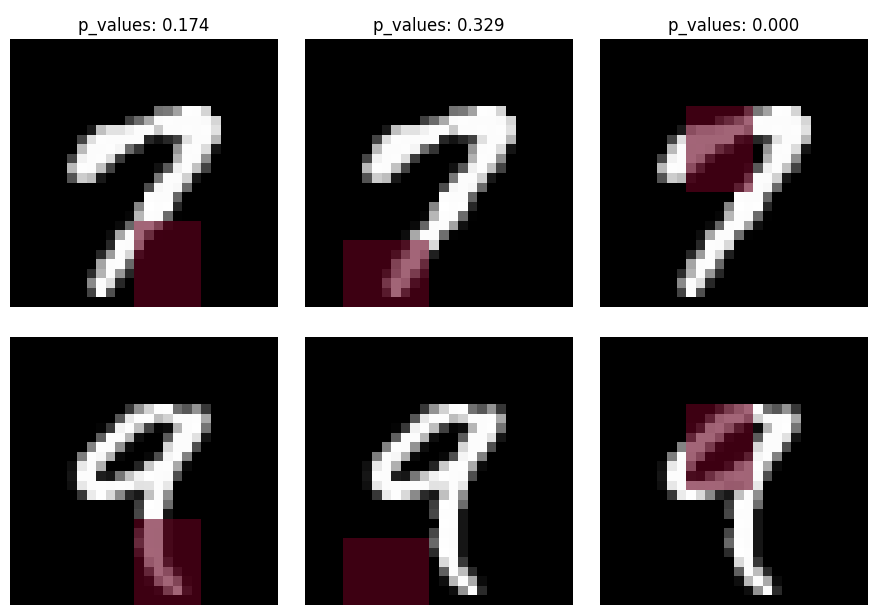}
\caption{The hypothesized regions (HRs) in Cases 1-3 for differentiating digits 7 and 9 in Section \ref{sec:app_mnist}. \textbf{Case 1:} a HR is $(19:28, 13:20)$, which indicates that $H_0$ is true; \textbf{Case 2:} a hypothesized region is $(21:28, 4:13)$, which indicates that $H_0$ is true; \textbf{Case 3:} a hypothesized region is $(7:16, 9:16)$, which indicates that $H_a$ is true. Note that the p-values in the top are given by the one-split test.}
\label{fig:app_demo}
\end{figure}

In this application, we consider three different types of masked regions, as displayed in Figure \ref{fig:app_demo}. 





To proceed, we specify the underlying model as the default convolution neural network (CNN) provided by Keras for the MNIST dataset. 
Finally, we apply the one-split test, the two-split tests, and their combined tests based on the data-adaptive tuning scheme with a significance level of $\alpha=0.05$.

\begin{table}[!ht]
  \caption{P-values of (combined) one-/two-split tests in the MNIST dataset. 
Significant p-values for testing feature irrelevance are underlined at a nominal level $\alpha = 0.05$.}
  \centering
\scalebox{1.0}{
\begin{tabular}{@{}cccccccccccccc@{}} \toprule
& & Test & \phantom{a} & & & \multicolumn{1}{c}{p-values (cases 1-3)} & \phantom{a} & \\
\midrule 
& & One-split && && (1.74e-1, 3.29e-1, \underline{1.37e-13})  \\
& & Two-split && && (9.59e-1, 5.69e-1, \underline{1.10e-05})  \\
& & Comb. one-split && && (3.85e-1, 1.00e-0, \underline{4.43e-18})  \\
& & Comb. two-split && && (5.44e-1, 1.92e-1, \underline{2.25e-09})  \\
\bottomrule
\end{tabular}}
\label{tab:app}
\end{table}

As suggested by Table \ref{tab:app}, the (combined) one-/two-split tests all fail to reject $H_0$ when it is true in Cases 1-2, but all reject $H_0$ in Case 3 when it is false.  
Overall, the test results confirm our intuition that the hypothesized regions in Cases 1-2 are visually indistinguishable, whereas that in Case 3 is visually discriminative, as illustrated in Figure \ref{fig:app_demo}. 

\subsection{Mechanisms of Action (MoA) prediction for new drugs}
\begin{figure}[ht]
  \centering
  \includegraphics[scale=.3]{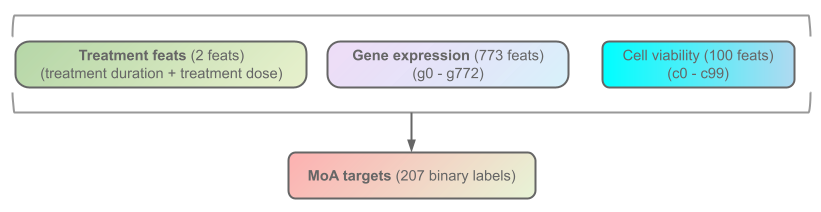}
  \caption{Features (treatment features, gene expression, and cell viability) and targets in Mechanisms of Action (MoA) dataset. Three cases with three different types of hypothesized features (HFs) are considered. \textbf{Case 1 (Treatment):} HFs are ``treatment duration'' and ``treatment dose''; \textbf{Case 2 (Gene):} HFs are ``g-0''-``g-772''; \textbf{Case 3 (Cell):} HFs are ``c-0''-``c-99''.  }
  \label{fig:app_MoA}
  \end{figure}

This subsection applies the proposed tests to examine the significance of ``treatment'', ``gene expression'', and ``cell viability'' to MoA prediction of new drugs. The dataset consists of 23814 drug-MoA annotation pairs with three types of features (``treatment'', ``gene expression'', and ``cell viability''), and 207 binary labels indicating multiple targets of MoA responses, as illustrated in Figure \ref{fig:app_MoA}. Specifically, ``treatment'' includes ``treatment duration'' (continuous) and ``treatment dose'' (categorical); ``gene expression'' and ``cell viability'' include 773 gene expression data (continuous), and 100 human cells' responses (continuous) to drugs \cite{corsello2020discovering, subramanian2017next}, respectively.

In this application, we consider the significance of those three types of feature sets, as displayed in Figure \ref{fig:app_MoA}. 

For implementation, we use TabNet \cite{arik2020tabnet} as the predictive model for our proposed tests. 
The results are summarized in Table \ref{tab:app_MoA}, which indicates that all tests fail to reject $H_0$ at $\alpha=0.05$ for ``gene expression'' features. For Cases 1 and 3, all tests consistently reject $H_0$, identify ``treatment'' and ``cell viability'' as significant features to MoA prediction.

\begin{table}[!ht]
  \caption{P-values of (combined) one-/two-split tests in the MoA prediction dataset.}
  \centering
\scalebox{1.}{
\begin{tabular}{@{}ccccp{4.5cm}cccc@{}} \toprule
&& Test & \phantom{a} & p-values (cases 1-3) \newline ('treatment', 'gene exp', 'cell viability')  \\
\midrule
& & One-split && (\underline{1.42e-2}, 1.34e-1, \underline{9.69e-4}) \\
& & Two-split && (\underline{2.17e-2}, 2.52e-1, \underline{3.19e-4}) \\
& & Comb. one-split && (\underline{4.72e-2}, 3.81e-1, \underline{1.02e-3}) \\
& & Comb. two-split && (\underline{1.13e-3}, 1.01e-1, \underline{1.20e-5}) \\
\bottomrule
\end{tabular}}
\label{tab:app_MoA}
\end{table}

\subsection{{Chest X-rays for pneumonia diagnosis}}
\label{sec:app_X}

This subsection illustrates the application of the proposed tests to chest X-ray images in a pneumonia diagnosis dataset \cite{kermany2018identifying}. 
This dataset consists of 5,863 X-ray images, each labeled as  ``Pneumonia'' or ``Normal''. To proceed, we crop an image to produce a version of the image that focuses on the lung fields, based on DeepXR. Then, we use a square cropping region to retain important areas containing parenchymal anatomy and retrocardiac anatomy. 

\begin{figure}[ht]
\centering
\includegraphics[scale=.24]{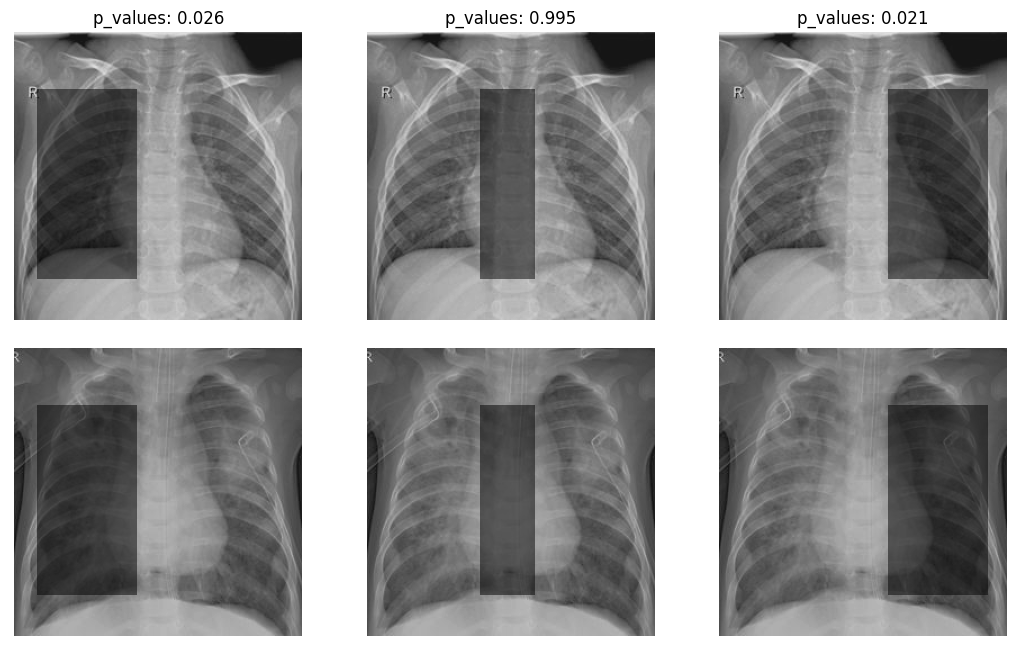}
\caption{The hypothesized regions (HRs) in Cases 1-3 for discriminating ``Normal'' (first row) versus ``Pneumonia'' (second row) X-ray images in Section \ref{sec:app_X}. \textbf{Case 1:} a HR is $(50:200, 20:110)$, for which $H_0$ is likely to be false; \textbf{Case 2:} a HR is $(50:200, 100:150)$, for which $H_0$ is likely to be true. \textbf{Case 3:} a HR is $(50:200, 150:240)$, for which $H_0$ is likely to be false. Note that the p-values in the top are
given  by the one-split test.}
\label{fig:app_demo_X}
\end{figure}

For implementation, we specify the learning model as a convolution neural network (CNN) and apply the one-split test, two-split test, and their combined tests based on the data-adaptive tuning scheme at a significance level of $\alpha=0.05$. Similarly, we also consider three different types of hypothesized regions, as displayed in Figure \ref{fig:app_demo_X}.

\begin{table}[!ht]
  \caption{P-values of the (combined) one-/two-split tests in the chest X-ray dataset.}
  \centering
\scalebox{1.}{
\begin{tabular}{@{}ccccp{5cm}cccc@{}} \toprule
&& Test & \phantom{a} & p-values (cases 1-3) \newline ('left lung', 'null region', 'right lung')  \\
\midrule 
& & One-split && (\underline{2.61e-2}, 9.95e-1, \underline{2.12e-2}) \\
& & Two-split && (2.12e-1, 5.61e-1, 6.51e-2) \\
& & Comb. one-split && (\underline{4.14e-2}, 6.35e-1, \underline{7.52e-2}) \\
& & Comb. two-split && (5.36e-2, 7.54e-1, 8.37e-2) \\
\bottomrule
\end{tabular}}
\label{tab:app_X}
\end{table}

As suggested by Table \ref{tab:app_X}, all tests fail to reject $H_0$ at $\alpha=0.05$ in Case 2 when $H_0$ is likely to be true. For Cases 1 and 3, only the (combined) one-split test rejects both the $H_0$, but other tests fail to do so when $H_0$ is likely to be false. In agreement with the earlier results, the one-split test seems more powerful to detect a discriminative region.

\subsection{Significance of keypoints to facial expression recognition}
This subsection examines the significance of five keypoints (left eye, right eye, eyes, nose, and mouth) on seven facial expressions: `angry', `disgust', `fear', `happy', `sad', `surprise', `neutral') on the FER2013 dataset, consisting of 48x48 pixel grayscale facial images. The facial images have been automatically registered.
For each facial image, an emotion label is provided as one of seven expressions.
Given a facial image, we produce the keypoints based on the existing facial landmark detection libraries \texttt{dlib} and \texttt{open-cv}. The primary goal is to deliver the significance of the keypoints to facial expression recognition. 

\begin{figure}[h!]
  \centering
  \includegraphics[scale=.25]{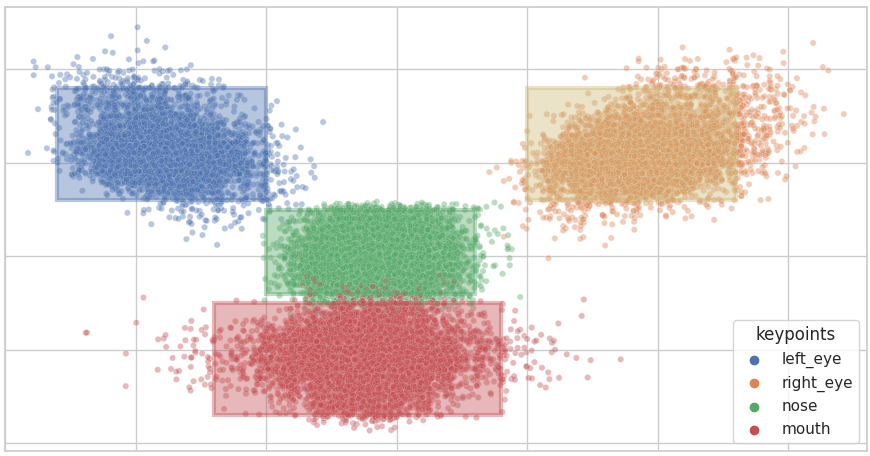}
  \caption{The scatter plot for the keypoints (left eye, right eye, nose, and mouth) in FER2013 facial expression recognition dataset, yielding that the hypothesized regions in Cases 1-4 cover the corresponding keypoints in most faces. }
  \label{fig:region}
\end{figure}

\begin{figure}[ht]
  \centering
  \includegraphics[scale=.30]{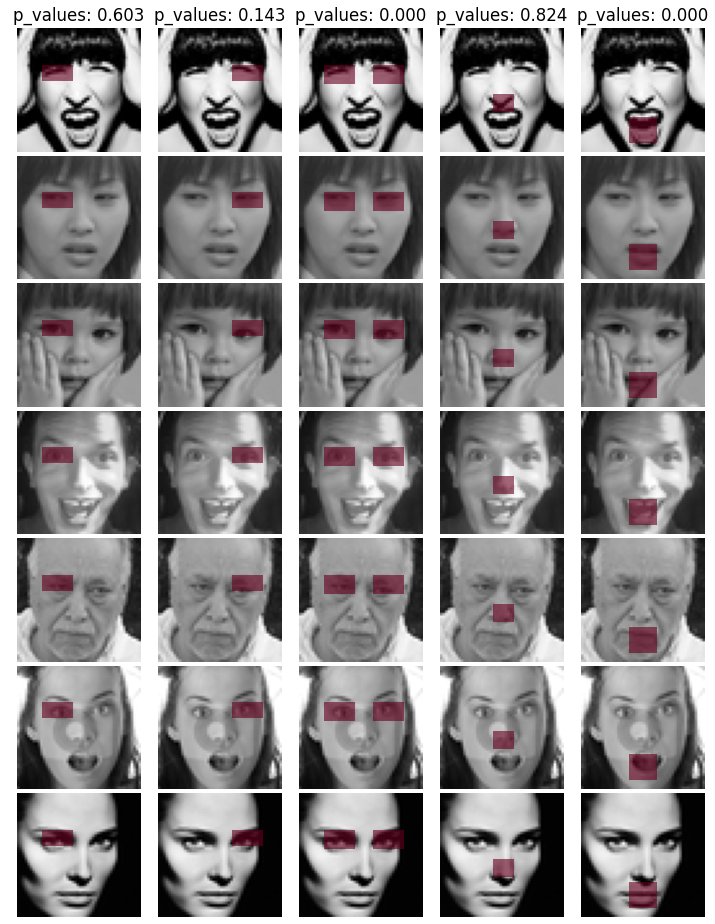}
  \caption{The hypothesized regions (HRs) in Cases 1-5 for discriminating seven facial expressions in rows (including `angry', `disgust', `fear', `happy', `sad', `surprise', `neutral'). 
  \textbf{Case 1 (Left eye):} a HR is (14:22, 9:22); 
  \textbf{Case 2 (Right eye):} a HR is (14:22, 28:41).
  \textbf{Case 3 (Eyes):} a HR is (14:22, 9:22 $\cup$ 28:41);
  \textbf{Case 4 (Nose):} a HR is (24:32, 20:29);
  \textbf{Case 5 (Mouth):} a HR is (34:45, 18:30).
  Note that the p-values in the top are given by the combined one-split test.}
  \label{fig:app_FER}
\end{figure}

After preprocessing, we obtain 11709 triples of images, labels, and keypoints. The scatter plot for the keypoints is provided in Figure \ref{fig:region}, from which we consider five different collections of hypothesized regions corresponding to five keypoints: left eye, right eye, eyes, nose, and mouth, respectively. The hypothesized regions based illustrative examples are displayed in Figure \ref{fig:app_FER}.

For implementation, we use the same VGG deep neural network and the same training hyperparameters as in \cite{khaireddin2021facial}. Note that the adopted VGG network in \cite{khaireddin2021facial} is one of the state-of-art facial expression recognition methods (Rank 4) in FER2013 \textit{papers-with-code} Leaderboard \cite{pramerdorfer2016facial}. 

\begin{table}[!ht]
  \caption{P-values of the (combined) one-/two-split tests in the FER2013 dataset based on five keypoints: left eye, right eye, eyes, nose, and mouth.}
  \centering
  \begin{tabular}{@{}ccp{6cm}cccc@{}} \toprule
    Test & \phantom{a} & p-values (cases 1-5) \newline ('left eye', 'right eye', 'eyes', 'nose', 'mouth')  \\
    \midrule 
     One-split && (1.58e-1, 3.55e-1, \underline{6.14e-3}, 6.89e-1, \underline{1.27e-3}) \\
     Two-split && (9.25e-2, 2.87e-1, \underline{1.75e-2}, 2.86e-1, \underline{3.46e-2}) \\
     Comb. one-split && (6.03e-1, 1.43e-1, \underline{7.88e-5}, 8.23e-1, \underline{1.09e-7}) \\
     Comb. two-split && (5.91e-2, 6.81e-2, \underline{4.42e-2}, 1.33e-1, \underline{2.29e-2}) \\
    \bottomrule
    \end{tabular}
\label{tab:app_FER}
\end{table}

As suggested by Table \ref{tab:app_FER}, all tests fail to reject $H_0$ in Cases 1, 2, and 4. For Cases 1-2, it is partly because the predictive information in the left/right eye is symmetrically leaked in the other eye. For Case 4, the result confirms the visual intuition that ``nose'' is not a discriminative keypoint to facial expression. For Cases 3 and 5, all tests consistently reject $H_0$, suggesting that ``eyes'' and ``mouth'' are discriminative regions, which are visually confirmed by illustrative samples in Figure \ref{fig:app_FER}. Note that the proposed tests are equally applicable to more substantial computer vision applications, for which the testing results could provide instructive information for visual sensor management and construction.

\subsection{Evaluating significance of localization in CIFAR100}
Note that the proposed methods are equally applicable to significance tests with instance adaptive hypothesized features. Therefore, they can be used to evaluate the effectiveness of discriminative localization methods, such as the class activation maps (CAM; \cite{zhou2016learning}) and Grad-CAM \cite{selvaraju2017grad}. In this subsection, we demonstrate a significant test in the CIFAR100 dataset based on adaptive hypothesized features localized by Grad-CAM. 

\begin{figure}[ht]
  \centering
  \includegraphics[scale=.22]{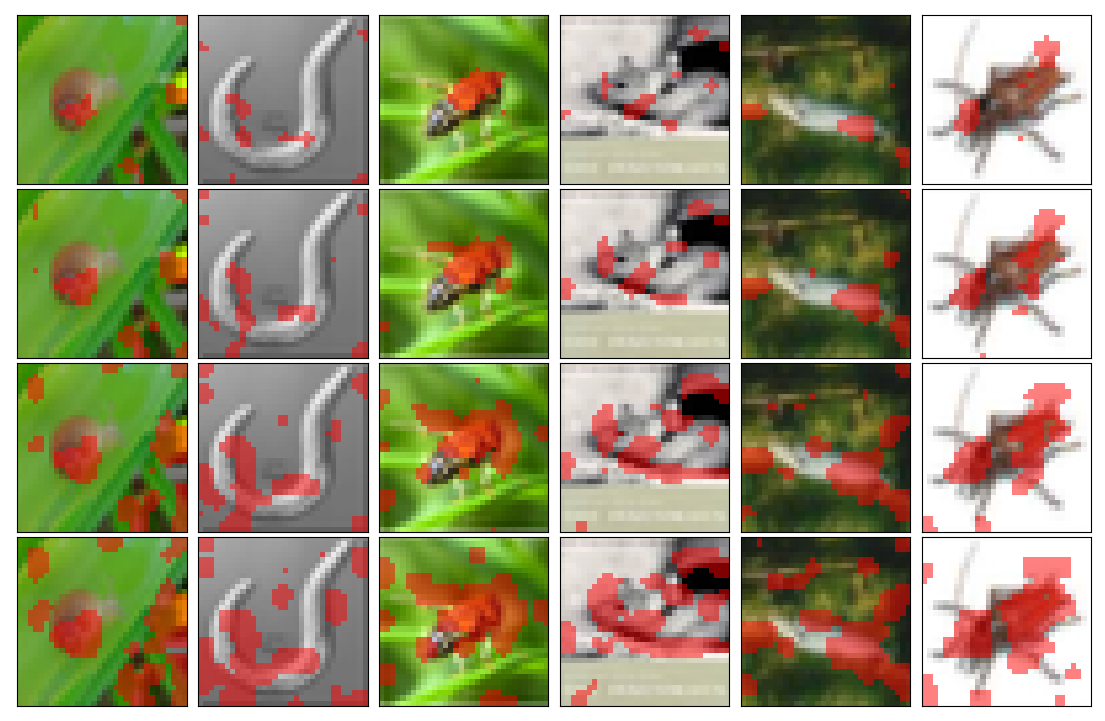}
  \caption{Demonstrative adaptive hypothesized regions in CIFAR100 dataset, localized by Grad-CAM.
\textbf{Cases 1-4}: the percentages of hypothesized features are 5\%, 10\%, 15\%, 30\% in rows from top to bottom, corresponding to the top important features ranked by Grad-CAM localization heatmaps.}
  \label{fig:app_CIFAR}
\end{figure}

Specifically, in the training set, we apply Grad-CAM to a fitted AlexNet to produce importance/heatmaps of features/pixels of all images, see 6 demonstrative examples in Figure \ref{fig:app_CIFAR}. Then, 4 cases of hypothesized tests are provided by taking top-5\%, top-10\%, top-15\%, top-30\% important features as hypothesized features. Next, the proposed one/two-split tests are conducted in the testing set with a ResNet50 network, and the resulting p-values are summarized in Table \ref{tab:app_CIFAR}.

Overall, the test results confirm our intuition, the inconsistent results in Cases 2 and 3 by one-split and two-split tests may be caused by the power loss of two-split tests. It is worth 
mentioning that the sequence of pairs (top important hypothesized/localized features, p-values) produced by the proposed tests can be an evaluation of the effectiveness of the localization method.

\begin{table}[!ht]
  \caption{P-values of the (combined) one-/two-split tests in the CIFAR100 dataset.
The percentages of hypothesized features are 5\%, 10\%, 15\%, 30\%, corresponding to the top important features 
ranked by Grad-CAM localization heatmaps.}
  \centering
  \begin{tabular}{@{}ccp{5cm}cccc@{}} \toprule
    Test & \phantom{a} & p-values (cases 1-4) \newline (Top-5\%, Top-10\%, Top-15\%, Top-30\%)  \\
    \midrule 
     One-split && (3.13e-1, \underline{6.44e-3}, \underline{5.33e-3}, \underline{4.25e-8}) \\
     Two-split && (9.04e-1, 2.58e-1, 7.90e-1, \underline{2.59e-4}) \\
     Comb. one-split && (5.56e-2, \underline{4.08e-3}, \underline{1.92e-5}, \underline{1.12e-7}) \\
     Comb. two-split && (5.81e-1, 1.59e-1, \underline{2.20e-2}, \underline{9.68e-5})  \\
    \bottomrule
    \end{tabular}
\label{tab:app_CIFAR}
\end{table}

\subsection{Significance of keywords in sentiment analysis}
This subsection examines the significance of keywords in sentiment classification based on the IMDB dataset \cite{maas2011learning}. This dataset provides 50,000 highly polar movie reviews for binary sentiment classification. We also obtain lists of positive, negative, and neutral opinion words from \cite{hu2004mining}. In this application,  we apply the proposed tests to examine the significance of positive/negative/neutral words contributing to sentiment analysis. For illustration, we report the results based on the top 350 frequent positive and negative-sentiment words and 350 randomly selected neutral-sentiment words in the IMDB dataset.

For implementation, we use a Bidirectional LSTM model as a prediction model for sentiment classification and apply the one-split/two-split tests and their combined tests based on the log-ratio splitting method at a significance level of $\alpha = 0.05$.

\begin{table}[!ht]
  \caption{P-values of the (combined) one-/two-split tests in the IMDB dataset with hypothesized 
features as: \textbf{Case 1}: the top 350 frequent positive words; \textbf{Case 2}: top 350 negative-sentiment words; \textbf{Case 3}: 350 randomly selected neutral-sentiment 
words.}
  \centering
  \begin{tabular}{@{}ccp{4.5cm}cccc@{}} \toprule
    Test & \phantom{a} & p-values (cases 1-3) \newline (positive, negative, neutral)  \\
    \midrule 
     One-split && (\underline{2.92e-2}, \underline{1.20e-3}, 3.37e-1) \\
     Two-split && (9.61e-2, 1.61e-1, 1.14e-1) \\
     Comb. one-split && (\underline{2.53e-5}, \underline{6.87e-3}, 1.29e-1) \\
     Comb. two-split && (2.98e-1, 2.24e-1, 6.20e-1)  \\
    \bottomrule
    \end{tabular}
\label{tab:app_IMDB}
\end{table}

Overall, the test results in Table \ref{tab:app_IMDB} confirm our intuition, where the positive and negative-sentiment words significantly contribute to sentiment analysis, but not neutral-sentiment words. Inconsistent results in Cases 1 and 2 by one-split and two-split tests may be caused by a power loss of the two-split test.

\section{Conclusions}

 This article proposes two novel risk-invariance tests, one-split and two-split tests, to assess the impact of 
a collection of hypothesized features on prediction.  Theoretically, we have established asymptotic null distributions of test statistics and their consistency in Type \rom{1}/\rom{2} errors. Numerically, we have demonstrated the utility of the proposed tests on simulated and real datasets. Next, we summarize some strengths and limitations of the proposed tests.

\textbf{Strengths.} (i) The proposed tests provide a practical inference tool for black-box models on complex data, which considerably relax assumptions in the existing literature. For example, CRT and HRT require a well-estimated conditional probability for features, which is often impractical. (ii) The proposed tests work for general risk-invariance testing on a collection of features of interest, which encompasses the conditional independence test when the log-likelihood loss is used. (iii) The proposed tests involve a limited number of model refitting, which suitable for
large-scale problems.

\textbf{Limitations.} (i) One-split/two-split tests split over the original dataset at the expense of reduced power or increased Type \rom{2} error. (ii) The log-ratio splitting scheme is conservative in that it prefers situations with a large estimation subset and a small inference subset.


\ifCLASSOPTIONcaptionsoff
  \newpage
\fi



\bibliographystyle{IEEEtran}
\bibliography{IEEEabrv,inf}
%





%

\begin{IEEEbiography}[{\includegraphics[width=1in,height=1.25in,clip,keepaspectratio]{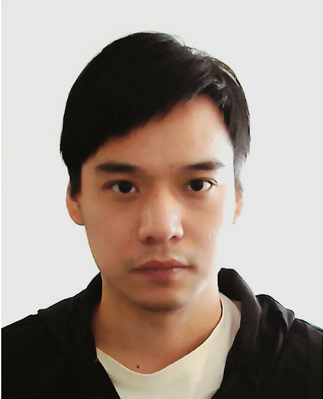}}]{Ben Dai} received his B.S. degree in Mathematics and Applied Mathematics from Hangzhou Dianzi University in 2015, and his Ph.D. degree in Data Science from the City University of Hong Kong in 2019. He is currently an Assistant Professor in Department of Statistics, The Chinese University of Hong Kong. His research interests include statistical machine learning, learning theory, statistical XAI, recommender systems, and deep learning.
\end{IEEEbiography}


\begin{IEEEbiography}[{\includegraphics[width=1in,height=1.25in,clip,keepaspectratio]{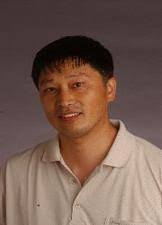}}]{Xiaotong Shen} received his B.S. degree in Mathematics from Peking University in 1985, and his Ph.D. degree in Statistics from the University of Chicago in 1991. He is the John Black Johnston Distinguished Professor in the University of Minnesota. His research interests include machine learning and data science, high-dimensional inference, nonparametric and semiparametric inference, causal graphical models, recommender systems and nonconvex minimization. He is a Fellow of the American Association for the Advancement of Science, the American Statistical Association, and the Institute of Mathematical Statistics.
\end{IEEEbiography}

\begin{IEEEbiography}[{\includegraphics[width=1in,height=1.25in,clip,keepaspectratio]{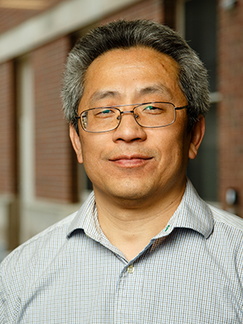}}]{Wei Pan} received his B.S. degrees in Computer Engineering and in Applied Mathematics from Tsinghua University in 1989, and his Ph.D. degree in Statistics from the University of Wisconsin-Madison in 1997. He is a Professor in Division of Biostatistics, the School of Public Health, the University of Minnesota. His research interests include statistical genetics, Bioinformatics and deep learning. He is a Fellow of the American Statistical Association and the Institute of Mathematical Statistics.
\end{IEEEbiography}




\end{document}